\documentclass[runningheads]{llncs}
\usepackage{graphicx}
\usepackage{tikz}
\usepackage{comment}
\usepackage{amsmath,amssymb} % define this before the line numbering.
\usepackage{color}
\usepackage[lofdepth,lotdepth]{subfig}

\usepackage[accsupp]{axessibility}  % Improves PDF readability for those with disabilities.

\newcommand{\tick}{\checkmark}
\usepackage[super]{nth}
\usepackage{booktabs}
\usepackage{tablefootnote}

\usepackage{bbm}
\newcommand{\mc}[1]{\multicolumn{1}{c}{#1}}
\newcommand{\dv}{DeepLabv$3$}

\newcommand{\hrn}{HRNet}

\newcommand{\upr}{UPerNet}

\newcommand{\tbf}[1]{\textbf{#1}}
\newcommand{\Lc}{\mathcal{L}}

\newcommand{\ca}[1]{\mathcal{#1}}
\newcommand{\enc}{f_{enc}}
\newcommand{\dec}{f_{seg}}
\newcommand{\til}[1]{\Tilde{#1}}

\usepackage[colorinlistoftodos,prependcaption,textsize=small]{todonotes}

\usepackage[ruled,vlined]{algorithm2e}
\usepackage{hyperref}

\begin{document}
% \linenumbers
\pagestyle{headings}
\mainmatter
\def\ECCVSubNumber{4181}  % Insert your submission number here

\title{Multi-scale and Cross-scale Contrastive Learning for Semantic Segmentation} % Replace with your title

% INITIAL SUBMISSION  % uncomment to anonymize
\begin{comment}
\titlerunning{ECCV-22 submission ID \ECCVSubNumber} 
\authorrunning{ECCV-22 submission ID \ECCVSubNumber} 
\author{Anonymous ECCV submission}
\institute{Paper ID \ECCVSubNumber}
\end{comment}
%******************

% SUPPLEMENTARY 
\begin{comment}
\titlerunning{Multi- and Cross-scale Contrastive Learning}

\author{}
\institute{}
\authorrunning{Pissas et al.}

\end{comment}

% \begin{comment}
\titlerunning{Multi- and Cross-scale Contrastive Learning}
\author{Theodoros Pissas\inst{1,2}  \and
Claudio S. Ravasio\inst{1,2} \and \\
Lyndon Da Cruz\inst{3,4}
\index{Da Cruz, Lyndon}\thanks{The last two authors contributed equally.} \and
Christos Bergeles\inst{2}\textsuperscript{*}} 
\authorrunning{Pissas et al.}
\institute{Wellcome/EPSRC Centre for Interventional and Surgical Sciences, University College London (UCL) \and
School of Biomedical Engineering \& Imaging Sciences, King's College London (KCL) \and
Moorfields Eye Hospital, London \and Institute of Ophthalmology, University College London \\
\email{rmaptpi@ucl.ac.uk}}

% \end{comment}

% comment to anonymize
% \begin{comment}

% \author{Theodoros Pissas\inst{1,2} \and
% Claudio S. Ravasio\inst{1,2} \and \\ 
% Lyndon Da Cruz\inst{3,4}\thanks{The last two authors contributed equally.} \and
% Christos Bergeles\inst{2}\textsuperscript{*}}
% %
% \authorrunning{Pissas, Ravasio, et al.}
% %
% \institute{Wellcome/EPSRC Centre for Interventional and Surgical Sciences, University College London (UCL) \and
% School of Biomedical Engineering \& Imaging Sciences, King's College London (KCL) \and
% Moorfields Eye Hospital, London \and Institute of Ophthalmology, University College London}

% \end{comment}

%******************
\maketitle
% \begin{comment}

\begin{abstract}
This work considers supervised contrastive learning for semantic segmentation. We apply contrastive learning to enhance the discriminative power of the multi-scale features extracted by semantic segmentation networks. Our key methodological insight is to leverage samples from the feature spaces emanating from multiple stages of a model's encoder itself requiring neither data augmentation nor online memory banks to obtain a diverse set of samples. To allow for such an extension we introduce an efficient and effective sampling process, that enables applying contrastive losses over the encoder's features at multiple scales. Furthermore, by first mapping the encoder's multi-scale representations to a common feature space, we instantiate a novel form of supervised local-global constraint by introducing cross-scale contrastive learning linking high-resolution local features to low-resolution global features. Combined, our multi-scale and cross-scale contrastive losses boost performance of various models (\dv, HRNet, OCRNet, UPerNet) with both CNN and Transformer backbones, when evaluated on $4$ diverse datasets from natural (Cityscapes, PascalContext, ADE20K) but also surgical (CaDIS) domains. Our code is available at \url{https://github.com/RViMLab/MS_CS_ContrSeg}.
\end{abstract}

% \end{comment}

%%%%% sections %%%%%%

\section{Introduction}
\label{sec:intro}
Supervised deep learning has driven remarkable progress in semantic segmentation, catalyzed by advances in convolutional network architecture design  and the availability of large-scale pixel-level annotated datasets. Regarding the former, the standard paradigm involves convolutional encoders \cite{ResNet,vgg,HRNet} to extract non-linear embeddings from images followed by a decoder that maps them to a task-specific output space. For semantic segmentation, the seminal work of fully convolutional networks \cite{FCN} demonstrated end-to-end learning of both these components by simply supervising the decoder with per-pixel classification losses. Further, considerable research has focused on designing inductive biases in convolutional architectures that enable complex, both local and global, visual relations across the input image to be encoded \cite{drn,deeplab,deeplabv3,OCR,HRNet}, leading to impressive results on challenging visual domains \cite{cts,ade20k}. 

%  Our work is orthogonal to these directions, as it proposes a contrastive learning loss for training such models.
\begin{figure}[t]
\begin{center}
\includegraphics[width=1\linewidth]{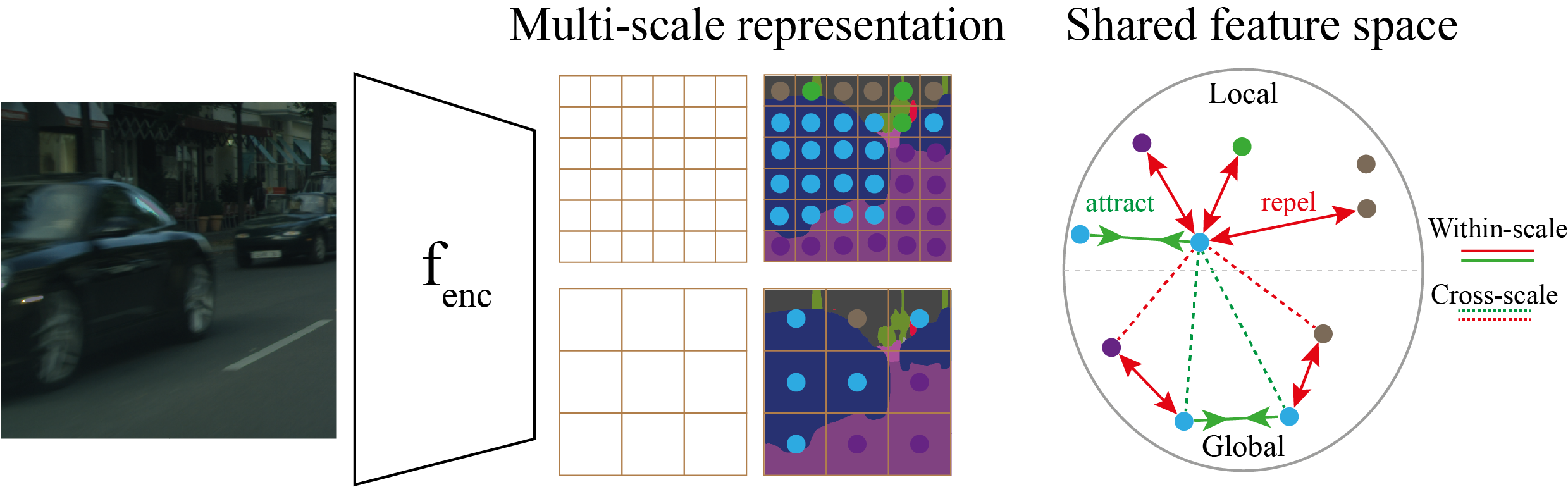}
\end{center}
\caption{\textbf{Key idea}: We leverage the multiscale representation provided by semantic segmentation encoders to propose a supervised-contrastive learning loss that is applied both at multiple scales and across them, within a shared feature space.
}
\label{fig:teaser}
\end{figure}

Recently, contrastive learning has been revisited as a tool for shaping network feature spaces under desired constraints over their samples, while avoiding representation collapse \cite{isola_align_uniformity}. It has achieved strong %empirical 
results in unsupervised representation learning, %as it has been 
and has established that contrastive pretraining over global \cite{moco,simclr,DetCo,DetCon} or dense \cite{DenseCL} features can give rise to encoders that are on par with their supervised counterparts when fine-tuned on downstream tasks. This progress has motivated the incorporation of contrastive loss functions and training methodologies for the tasks of supervised classification and segmentation.

For the latter, recent works have shown that dense supervised contrastive learning applied over the final encoder layer can boost semantic segmentation performance when labelled examples are scarce \cite{label_efficient_contrastive}, or in semi-supervised learning \cite{contr_semi_mem}. When using the full datasets, the method of \cite{exploring_contrastive} showed significant overall gains when enhanced with a class-wise memory bank maintaining a large diverse set of features during training. While recognizing and delineating objects at multiple scales is essential for segmentation, these methods directly regularize only the feature space of the final encoder layer thus operating at a single scale. 

We instead apply multi-scale contrastive learning at multiple model layers. This direct feature-space supervision on early convolution/attention layers, usually learnt by back-propagating gradients all the way through many layers starting from the output space, can allow them to capture more complex  relations between image regions, complementary to the usual function of early layers as local texture or geometry feature extractors. Intuitively, we treat the network as a function that maps an image to a multi-scale representation, distributed across its different stages, and endow it with class-aware, feature-space constraints. 

A second central element of contrastive approaches is the sampling of positive and negative samples to respectively attract and repel same-class embeddings. The usual sample generation mechanisms for unsupervised learning are augmentations, which as extensively discussed in \cite{simclr,moco,not_be_cont}, is a non-trivial and lengthy tuning step. Similarly, maintaining a class-wise memory bank, while providing a large set of examples, introduces hyper-parameters such as its size, update frequency and sample selection heuristic that can be dataset dependent \cite{exploring_contrastive}. 

We propose a simpler%, hyper-parameter-free
alternative to those options, by collecting samples from the feature spaces of multiple layers. Essentially, we leverage internal, intermediate information from the encoder that is available simply through the feedforward step of the network without the need to resort to external steps such as data augmentation or online storing of samples via memory banks. 
 Specifically, to find diverse \textit{views} of the data, we sample them from within the model's multi-scale feature spaces and link them both within and across scales (Fig\ \ref{fig:teaser}). Conclusively, our contributions are the following:  
\begin{itemize}
    \item A batch-level hyper-parameter-free anchor sampling process that balances the contributions from frequent and rare classes while allowing efficient application of multiple contrastive loss terms.  
    \item The introduction of supervised contrastive loss terms at multiple scales, and a novel cross-scale loss that enforces local-global consistency between high-resolution local and low-resolution global features emanating from different stages of the encoder. 
    \item A model-agnostic overall method that can be successfully applied to a variety of strong baselines and state-of-the-art methods, comprising both transformer and CNN-based backbones, on $4$ challenging datasets leading to improved performance.
\end{itemize}

\noindent Notably, on ADE20K, our method improves OCRNet \cite{OCR} by $\mathbf{2.3\%}$ and \upr\ with Swin-T and Swin-S, by $\mathbf{1.4\%}$ and $\mathbf{1.3\%}$, respectively (for single scale evaluation).  Further, on Cityscapes, our approach achieves an improvement of $\mathbf{1.1\%}$ for both \hrn\ \cite{HRNet} and UPerNet with ResNet-$101$. Finally, on a challenging surgical scene understanding dataset, CaDIS, we outperform the state-of-art \cite{what_matters}, especially improving rare classes by $\mathbf{1.2\%}$.

\section{Related works}
\label{sec:related}

We outline connections between several research areas and our method
and discuss its differences with existing methods.

\noindent \textbf{Context aggregation for semantic segmentation:}
An intrinsic property of convolutional encoders is that information is aggregated from each pixel's local neighborhood, the size of which can be expanded at the sacrifice of spatial feature resolution. The latter, however, is crucial for segmentation. The simple and effective aggregation approach of \cite{FCN} %demonstrated that a simple aggregation approach is to resize and sum multiscale feature maps it 
underpinned extensive research into more sophisticated context-aggregation mechanisms such as dilated-convolution-based approaches \cite{deeplab,drn,Deeplabv3plus,deeplabv3}, attention \cite{OCR,GatedSCN}, feature pyramid networks \cite{UPerNet}, or maintaining a high resolution representation throughout the network \cite{HRNet}. Recently, transformer and hybrid architectures have also been proposed. By design, these enable long-range context aggregation at the cost of increased computational requirements for training and inference \cite{transformer1,transformer2}.
Broadly, these approaches only support information aggregation from within a single input image. On the contrary, our method links features from different images during training, while being compatible with any architecture with hierarchical structure.  

\noindent \textbf{Contrastive learning} is a feature learning paradigm under which, given a known assignment of samples into classes, the objective is to minimize the distance between samples of the same class while maximizing the distance between samples of different classes. From an information-theoretic perspective, minimizing an instantiation of this objective, termed the \textit{InfoNce} loss \cite{InfoNCE}, maximizes a lower bound of the mutual information (MI) between same class features. The same loss was used to learn a pretext task for unsupervised representation learning leading to impressive downstream task results \cite{InstanceDiscr,moco,simclr,DenseCL,DetCo}, while in \cite{SupervisedContrastive} it was shown to perform on par with standard cross-entropy for classification. 

\noindent \textbf{Supervised contrastive learning for segmentation:} Concurrently with this work, improvements on strong baselines were demonstrated for segmentation by employing a \textbf{single} supervised contrastive loss term as an auxiliary loss applied after the inner-most layer of the encoder \cite{exploring_contrastive,region_aware_contrast}. We instead explore %the extension to 
\textbf{multiple} contrastive loss terms computed over different encoder layers, directly regularizing their feature spaces. Both \cite{exploring_contrastive,region_aware_contrast} %works 
employ a memory bank used to maintain an extended set of pixel-embeddings or class-wise averaged region-embeddings. Instead, we opt for a simpler memory-less approach that avoids the need to tune memory size and memory update frequency, and instead collect samples from within and across different encoder layers. Further, while \cite{exploring_contrastive,region_aware_contrast} focused on ResNet and HRNet, we demonstrate effective application of our method on a wider set of architectures and backbones, also exploring UPerNet and transformers. Finally, in \cite{label_efficient_contrastive}, a similar contrastive loss term was used as a pretraining objective for \dv\ leading to significant gains in performance when labelled examples are scarce. However, it provided small benefits when using the complete datasets, and required long ($300-600$K steps) $2$-phase training schedules, which is close to $3$-$6\times$ the training steps for our approach. Notably, in \cite{label_efficient_contrastive}, contrasted feature maps had to spatially downsampled, for efficiency, which we circumvent by employing a balanced sampling approach.

\noindent \textbf{Local-global representation learning:}
Contrastive learning with local and global representations of images has been studied for unsupervised pretraining. \cite{MI_bengio} proposes training an encoder by maximizing the MI between local and global representations to force the latter to compactly describe the shared information across the former. The method of \cite{sentence_local_global} is similar; the maximization of MI between features encoding different local windows of a sentence, and global sentence embeddings, is used as a pretraining objective in natural language processing leading to improved downstream transfer. Moreover, in \cite{MI_views}, the InfoNCE loss is optimized over local and global features computed for two augmented views of an image, while in \cite{medical_local_global} it is applied on medical images separately over global features, and local features. In the latter, computation of the local loss term assumes the presence of common anatomical structures in medical scans to a relatively fixed position, %This allows the derivation of positive/negative samples for contrastive learning based on ``hard-coded'' pixel coordinates, 
an assumption  invalid for datasets with diverse structures and scenes. There, strict prior knowledge on the location of an object/region is unrealistic. Finally, in \cite{DetCo} with the goal of pretraining for object detection, multiple InfoNCE objectives are optimized: globally, by extracting a feature vector for the whole image, locally by extracting multiple feature vectors each describing a single random patch of the same image, and across local-global levels by forcing whole-image and patch-level features of the same image to be similar. 

Our approach is supervised, allowing to directly align the negative/positive assignment with the task of segmentation. We do not employ augmentation to obtain views of the data as in \cite{MI_views,medical_local_global}, as tuning it is laborious \cite{simclr,not_be_cont}. Instead, we leverage each dataset's spatial class distribution as a source for diverse samples. Finally, our local-global loss term is better aligned to segmentation than \cite{DetCo}: instead of enforcing scale invariance of globally pooled encoder features, our loss is computed over dense features from multiple layers.% and enforces part-to-whole consistency. 

\section{Method}
\label{sec:method}
We now provide an introduction to the InfoNCE loss function and the proposed multi-scale and cross-scale losses, as well as a sampling procedure to perform balanced and computationally feasible contrastive learning.

\subsection{Preliminaries}
\label{subsec:notation}

Let $I \in \mathbb{R}^{H \times W \times 3}$ be an image, with $H$ its height and $W$ its width. Semantic segmentation frameworks usually comprise a backbone encoder and a segmentation head. %, denoted as the encoder 
The backbone encoder, $\enc : \mathbb{R}^{H \times W \times 3} \mapsto \mathbb{R}^{h \times w \times c}$, maps the input image to a $c$-dimensional feature space. The segmentation head, $\dec : \mathbb{R}^{h \times w \times c} \mapsto [ \,0,1 ]^{H \times W \times N_c}$, decodes the features $F=f_{enc}(I)$ to produce the output per-pixel segmentation $\hat{Y}=\dec(F)$. These two modules are usually trained with a \textit{cross-entropy} loss $\Lc_{ce}(\hat{Y},Y)$, where $Y$ denotes the ground truth per-pixel labels. Both $\enc$, $\dec$ can be learned end-to-end with only supervision of $\dec$. Under this training paradigm, the learning signal is restricted to unary classification errors.

An extension to this paradigm is to produce gradients that consider the distance of encoded image regions relative to other regions (not necessarily from the same image) in feature space.
To directly (rather than implicitly via the decoder's gradients) shape the encoder's latent space, a supervised contrastive loss term \cite{SupervisedContrastive} can be used. The loss forces features from image regions belonging to the same (different) class to be pushed close (apart). In this work, to identify the classes in each image, we use the labels downsampled to the spatial dimensions of the feature space, which we denote by $\til{Y}$. Given this class assignment, the InfoNCE \cite{InfoNCE} loss can be computed over a set of feature vectors, which is usually termed \textit{anchor} or \textit{query} set; we denote this set by $\ca{A}$. For $\forall z_i \in \ca{A}$, there exist the sets of positives and negatives denoted by $\ca{P}_i$ and $\ca{N}_i$ respectively, which in the supervised setting that we examine, are identified according to the labels. The availability of dense rather than global features and labels allows a simple way to identify positive samples, referred also as ``views'', without the need to craft a set of appearance and geometric perturbations as done in other unsupervised \cite{simclr,moco} and supervised approaches \cite{label_efficient_contrastive}. Thus, instead of requiring a dataset-specific data augmentation pipeline, we exploit the natural occurrences of same or different class-pixels across the scene and across different images.  

\begin{equation}
\label{eq:infoNCE}
\Lc_{c} (\ca{A}) = \frac{1}{\mid \ca{A} \mid}\sum_{i \in \ca{A}} \frac{1}{\mid \ca{P}(i) \mid}  \sum_{j \in \ca{P}(i)} \Lc(z_i, z_j) ,
\end{equation}
where 
\begin{equation}
\Lc(z_i, z_j) = - \log \frac{\exp{(z_i\cdot z_j / \tau)}}{\exp{(z_i\cdot z_j / \tau)} + \sum_{n \in \ca{N}(i)} \exp{(z_i\cdot z_n / \tau)}}
\end{equation}

\noindent This formulation is identical to the supervised contrastive loss for classification proposed in \cite{SupervisedContrastive}. There, however, the choices of $\ca{P}_i$ and $\ca{N}_i$ are straightforward as each image in a batch has only a single global label. Crucially, as is standard practice in contrastive learning over convolutional feature maps \cite{moco,simclr}, the loss is not directly computed over the encoder features $F$, but rather over a non-linear projection using a small FCN $f_{proj}: \mathbb{R}^{h \times w \times c} \mapsto \mathbb{R}^{h \times w \times d}$ such that $Z=f_{proj}(F)$. For the rest, $z_i$ refers to a $d$-dimensional feature vector from the $i$-th spatial position of $Z$. %$Z \in \mathbb{R}^{h \cdot w \cdot d}$
We now motivate and describe the proposed anchor sampling process and the multi- and cross-scale loss terms.

\subsection{Fully-dense contrastive learning}
\label{subsec:anchors_negs_pos}
\label{subsec:dcv0}

\noindent In the general case, $\ca{A}$ can consist of all feature vectors from all spatial positions of $F$, the set of which is hereby denoted as $\Omega_F$. For each element $z_i$ of $\ca{A}$ and knowing the downsampled labels $\til{Y}$, we select as $\ca{N}_i=\{z_j \in\ca{A} : j \neq i, \til{Y}(j)\neq \til{Y}(i) \}$ and
$\ca{P}_i=\{ z_j \in\ca{A} : j\neq i, \til{Y}(j)=\til{Y}(i) \}$. The computational complexity of this operation is quadratic in the spatial dimensions of the feature vector, i.e., $\ca{O}(h^2 w^2)$. As most semantic segmentation methods, e.g. \cite{Deeplabv3plus,HRNet,OCR}, require a small output stride for $\enc$, it can become prohibitively expensive. Additionally, the quadratic complexity of this choice becomes even more prohibitive when introducing contrastive losses at multiple layers. Further, it is a well studied property of contrastive losses that the number \cite{moco,simclr} and the hardness \cite{contrastive_theory,kalantidis_mixing,InfoNCE} of the negatives affect the learned representations. Therefore, minimizing (\ref{eq:infoNCE}) over $\Omega_F$ can become trivial due to the consideration of many easy samples. 

\subsection{Anchor-set sampling}
\label{subsec:sampling}
\begin{figure}[t]
\begin{center}
\includegraphics[width=.9\linewidth]{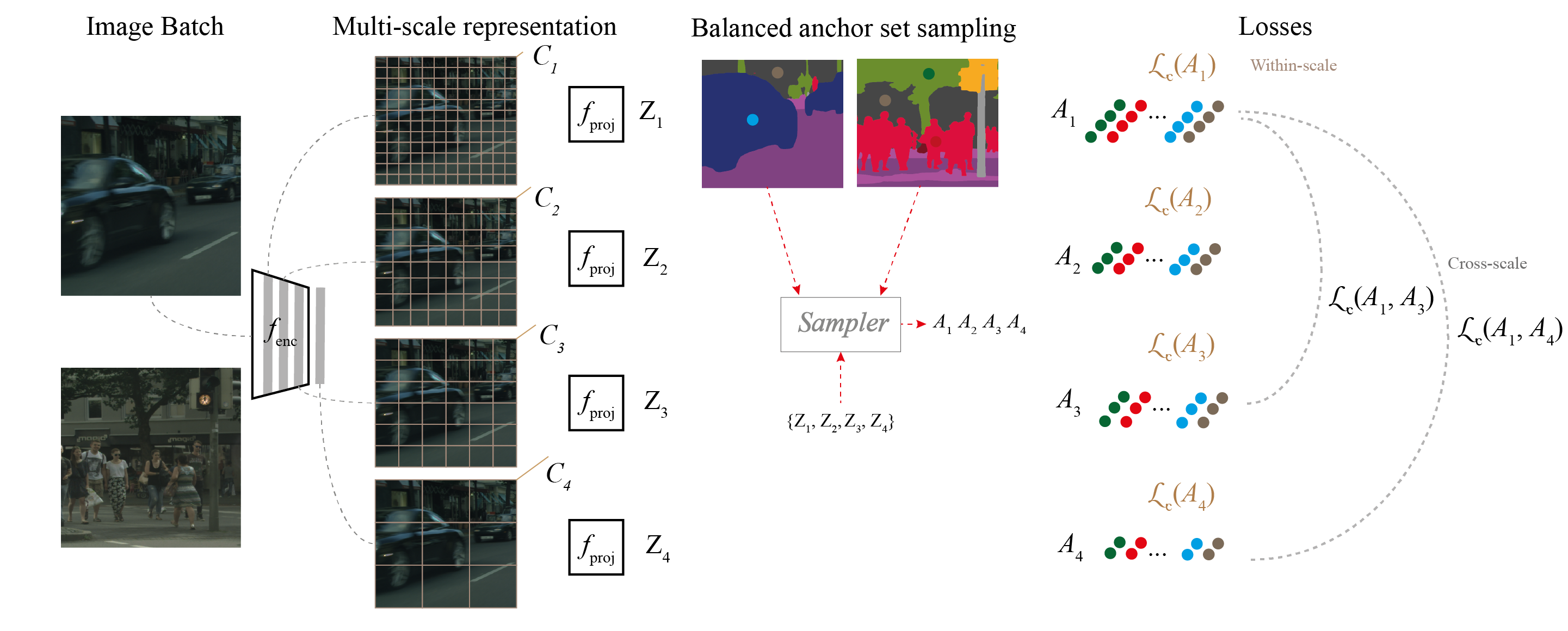}
\end{center}
\caption{\textbf{Method overview:} \textbf{(1)} An input batch is mapped to a multi-scale representation $F_1, ... F_4$, each having varying spatial dimensions and channels $C_1,..., C_4$, using the convolutional encoder. \textbf{(2)} Each dense feature is projected using a separate $f_{proj}$ that preserves spatial dimensions but maps all features to a common $d$-dimensional space. \textbf{(3)} Each projected feature map $Z_s$ undergoes a label-based balanced sampling process (Sec. \ref{subsec:anchors_negs_pos}) to produce an anchor set $\ca{A}_s$. \textbf{(4)} The feature vectors in $\ca{A}$ and their class assignments are used to calculate the contrastive loss of (\ref{eq:infoNCE}) within each scale. Further, pairs of anchor sets from two different scales are utilized to compute a cross-scale loss (Sec. \ref{subsec:dcv2_cs}). }
\label{fig:overview}
\end{figure} %< bloody latex and its heuristics for figure placement

%\noindent\textbf{Anchor-set sampling} 
An alternative is to sample from $\Omega_F$ while maintaining a balanced number of feature vectors from each present class. We generate $\ca{A}$ across a batch of images, comparably to \cite{exploring_contrastive}. The intuition behind this is to further expand the diversity of samples considered by not restricting semantic relations between samples to span a single image. Rather, we allow samples to extend across different images. 

% The proposed sampling has
% two advantages: provides computational efficiency (see B),
% and balanced contributions of classes to the loss. Setting
% sampled anchors per class to those of the class with the least
% samples in the batch preserves the effect of small pixel-size
% classes to the loss. This is supported on both datasets we
% tested on: 1) on CTS, the iIoU, which reweights IoU by object size, is significantly increased (see A); 2) on CADIS,
% mIoU of rare classes is boosted (Tables 3,4).

Specifically, we sample $K$ anchors per class present in the batch, equally divided among instances of that class. The sets of positives and negatives of each anchor are populated with elements of $\ca{A}$. This sampling process can be described as $\ca{A} \thicksim \{i \in \cup_{b=1}^{B} \Omega_F^{(b)} \}$, such that $\ca{A}$ has at most $A_{max}$ samples, with $B$ being the batch size. Importantly and contrary to \cite{exploring_contrastive,label_efficient_contrastive}, $K$ is selected on-the-fly rather than as a hyper-parameter, and is the number of samples from the class with the least occurrences in the batch. This is motivated by the observation that classes of small objects or regions will only occupy a small fraction of $\Omega_F$, even more so due to the spatial stride of network encoders. This heuristic enables a balanced contribution of semantic classes in the loss and removes the requirement for tuning $K$.
Additionally, it reduces the computational cost of each contrastive term enabling the multi-scale and cross-scale extensions described next. 
%As a preliminary experiment, shown in Table \ref{tab:ablations_sampling} \todo[inline]{add table re sampling}, we validated that our approach performs slightly better than the one examined in the implementation of \cite{exploring_contrastive} where the number of samples per class $K$ is chosen offline. 
% \begin{align}
% \mid \ca{A} \mid \leq A_{max} \\ \sum_{b=1}^{B}\sum_{i \in \ca{A}} \mathbbm{1}[\til{Y}^{b}(i)=y]=K  \quad \forall \quad  y \quad \text{in batch}.
% \end{align}
 
\subsection{Multi-scale contrastive loss}
\label{subsec:dcv2_ms}

Having obtained an efficient way to compute the loss of (\ref{eq:infoNCE}), we extend it to multiple scales. This extension regularizes the feature space of different network layers, by pushing same-class features closer, and maximizes the MI of features and their semantic labels. Importantly, while applying a pixel-space classification loss would also achieve the latter, the way we generate the anchor set allows us to attract features of the same class \textit{from across different images}. 

The hierarchical design of convolutional (ResNet \cite{ResNet}, HRNet \cite{HRNet}) or transformer (Swin \cite{swin}) encoders provides a natural interface over which our loss can be applied. The stages followed are also outlined in Figure \ref{fig:overview}. We independently generate $\ca{A}_{s}$ (Sec.~\ref{subsec:sampling}) using the projection of features $F_{s}$, $Z_s$ and labels $\til{Y}_{s}$ at each scale $s$. The overall loss is computed as a weighted sum across scales:
\begin{equation}
\label{eq:cont_ms}
\Lc_{cms} = \sum_{s=1}^{S} w_{s}  \Lc_{c}(\ca{A}_{s})
\end{equation}
\noindent where the weights $w_{s}$ control the contribution of each scale in the overall loss, and $S$ is the number of different scales. As described in Section \ref{subsec:sampling} generating $\ca{A}_{s}$ involves randomization, thus it is ensured that the image regions involved in computing the above loss at each scale are independent.   

\subsection{Cross-scale contrastive loss}
\label{subsec:dcv2_cs}

We further push same-class features from across different scales closer together. Specifically, we push high-resolution local features to be close to lower resolution global features. Given that global features encapsulate high level semantic concepts of the encoded image, guided by $Lc_{ce}$, we require that local features also encode those concepts by forcing them to lie close by in the projector's feature space. Intuitively, this enables local features describing parts of objects/regions to be predicative of their global structure of the object and vice versa. 

Importantly, we note that directly requiring that the features be similar across scales would be very hard to satisfy without causing collapse. The use of separate small non-linear convolutional projector at each scale (Figure \ref{fig:overview}) provides a compromise between a hard contrastive constraint on the encoder's features and the lower dimensional common space spanned by $f_{proj}$ wherein the cross-scale loss is calculated. Therefore, we compute it using the independently generated anchor sets $\ca{A}_s$ and $\ca{A}_{s'}$, from scales $s$ and $s'$:

\begin{equation}
\label{eq:cont_cs}
\Lc_{ccs} = \sum_{(s,s')} w_{s,s'}  \Lc_{c}(\ca{A}_{s}, A_{s'})
\end{equation}

\noindent Negatives and positives for samples of one scale are collected from the anchor set of the other. Concretely, for each element $z_i \in \ca{A}_s$ we specify $\ca{N}_i=\{ z_j \in\ca{A}_{s'}: j \neq i, \til{Y}_{s'}(j)\neq \til{Y}_{s}(i) \}$ and $\ca{P}_i=\{ z_j \in\ca{A}_{s'} : j\neq i | j,  \til{Y}_{s'}(j)=\til{Y}_{s}(i) \}$. Finally, the gradients derived from this loss are backpropagated to both involved anchor sets thus instantiating a form of bidirectional local-global consistency for learning the encoder. Combining all the above terms, our complete objective is:
\begin{equation}
\label{eq:total_loss}
\Lc_{total} = \Lc_{ce} + \lambda_{cms} * \Lc_{cms} + \lambda_{ccs} * \Lc_{ccs} 
\end{equation}

\begin{table}[ht]
\centering
\caption{(a),(c): Ablations for each component of our contrastive loss on \textbf{Cityscapes val} and \textbf{ADE20K}, respectively, reporting the mean and std deviation of each variant across 4 random seeds. (b), (d): Comparisons with auxiliary losses on \textbf{Cityscapes} (results taken from \cite{exploring_contrastive}) and \textbf{ADE20K}, respectively.}\label{tab:abl_losses_comp_aux}
\subfloat[a][]{  
  \resizebox{72mm}{!}{
  \begin{tabular}{*7c}
    \toprule
    \multicolumn{2}{c}{Network}   & \multicolumn{4}{c}{Loss} & \mc{mIoU (ss)}  \\
    \cmidrule(r){1-2}  \cmidrule(){3-6}  \cmidrule(l){7-7} 
    \tbf{Model}      & \tbf{Backbone}     & $\Lc_{c}$  & $\Lc_{cms}$  & $\Lc_{ccs}$ & Scale pairs  & mean$\pm$std \\
    \midrule
    \mc{\hrn}   & \mc{HR$48$v$2$}        & \mc{}         & \mc{}            & \mc{}        &  -   & \mc{$79.1 \pm 0.2$} \\ 
    \mc{\hrn}   & \mc{HR$48$v$2$}        & \mc{\tick}    & \mc{}            & \mc{}        &  -   & \mc{$80.0 \pm 0.1$} \\ 
    \mc{\hrn}   & \mc{HR$48$v$2$}        & \mc{}         & \mc{\tick}       & \mc{}        &  -   & \mc{$80.7 \pm 0.2$} \\ 
    \mc{\hrn}   & \mc{HR$48$v$2$}        & \mc{}         & \mc{\tick}       & \mc{\tick}   &  1   & \mc{$81.2 \pm 0.1$} \\ 
    % s4-s32
    \mc{\hrn}   & \mc{HR$48$v$2$}        & \mc{}         & \mc{\tick}       & \mc{\tick}   &  2   & \mc{$\mathbf{81.4\pm 0.1}$} \\ 
    \bottomrule
  \end{tabular}}
  } % subfloat
\qquad
\subfloat[b][]{
\resizebox{36mm}{!}{
  \begin{tabular}{*3c}
    \toprule
    \mc{Model}   &  \mc{Loss}                                   & \mc{mIoU (ss)}  \\
    \midrule
    \mc{\hrn}                        & CE                                           & \mc{79.1} \\
    \mc{\hrn}                        & \mc{AFF} \cite{AFF}                          & \mc{78.7} \\ 
    \mc{\hrn}                        & \mc{RMI} \cite{RMI}                          & \mc{79.9} \\ 
    \mc{\hrn}                        & \mc{Lovasz} \cite{Lovasz}                    & \mc{80.3} \\
    \mc{\hrn}                        & \mc{ours}                                    & \mc{\textbf{81.5}} \\
    \bottomrule
  \end{tabular}}}
\qquad
\subfloat[c][]{
\resizebox{72mm}{!}{
\begin{tabular}{*7c}
    \toprule
    \multicolumn{2}{c}{Network}   & \multicolumn{4}{c}{Loss} & \mc{mIoU (ss)}  \\
    \cmidrule(r){1-2}  \cmidrule(){3-6}  \cmidrule(l){7-7} 
    \tbf{Model}      & \tbf{Backbone}     & $\Lc_{c}$  & $\Lc_{cms}$  & $\Lc_{ccs}$ & Scale pairs  & mean$\pm$std \\
    \midrule
    \mc{UPerNet}   & \mc{Swin-T}        & \mc{}         & \mc{}            & \mc{}        &  -   & \mc{$44.5 \pm 0.4$} \\ 
   \mc{UPerNet}   & \mc{Swin-T}        & \mc{}         & \mc{\tick}       & \mc{}        &  -   & \mc{$45.1 \pm 0.2$} \\ 
    \mc{UPerNet}   & \mc{Swin-T}        & \mc{}         & \mc{\tick}       & \mc{\tick}   &  2   & \mc{$\mathbf{45.8 \pm 0.1}$} \\ 
    \bottomrule
\end{tabular}}}
\qquad
\subfloat[d][]{
\resizebox{34mm}{!}{
  \begin{tabular}{*3c}
    \toprule
    \mc{Model}   &  \mc{Loss}                                   & \mc{mIoU (ss)}  \\
    \midrule
    \mc{OCRNet}  & CE                                           & \mc{44.5} \\
    \mc{OCRNet}  & \mc{Lovasz}                                  & \mc{44.7} \\
    \mc{OCRNet}  & \mc{ours}                                    & \mc{\textbf{46.8}} \\
    \midrule
    \mc{SwinT}  & CE                                           & \mc{44.5} \\
    \mc{SwinT}  & \mc{Lovasz}                                  & \mc{45.2} \\
    \mc{SwinT}  & \mc{ours}                                    & \mc{\textbf{45.9}} \\
    \bottomrule
  \end{tabular}}}
\qquad
\end{table} % ablations on loss components 

\section{Experiments}
\label{sec:experiments}
%We now provide details on the experiments we use to evaluate our method.
\subsection{Datasets}
\label{subsubsec:datasets}
We benchmark our approach on the following challenging datasets from natural and surgical image domains using the mIoU as the main performance metric:

\noindent \tbf{ADE20K} \cite{ade20k} comprises $20210$ and $2000$ and train and val images, respectively, capturing $150$ semantic classes from natural scenes.  

\noindent \tbf{Cityscapes} \cite{cts} consists of $5000$ images of urban scenes on which $19$ classes are pixel-level labelled. We train, and evaluate, on the \textit{train}, and \textit{val} sets, respectively. We also report performance on the server-withheld \textit{test} set. 

\noindent \tbf{Pascal-Context} \cite{PascalContext} comprises $4998$ and $5105$ and train and val images, respectively, capturing $59$ classes from natural scenes.  

\noindent \tbf{CaDIS} \cite{cadis} comprises $25$ cataract surgery videos and $4671$ pixel-level annotated frames with labels for \textit{anatomies}, \textit{instruments} and \textit{miscellaneous objects}. We experiment on tasks $2$ and $3$ defined in \cite{cadis}, comprising %$9$ surgical tool classes, $6$ anatomies and $2$ miscellaneous objects for a total of 
$17$, %The more challenging task $3$ pushes instrument recognition granularity even further by considering the handles of certain instruments as separate classes, and has 
$25$ classes respectively. We follow \cite{what_matters} and also report the average of the per-class IoUs for anatomies, surgical tools and rare classes (present in less than $15\%$ of the images).

\begin{figure}[t]
\begin{center}
\includegraphics[width=.99\linewidth]{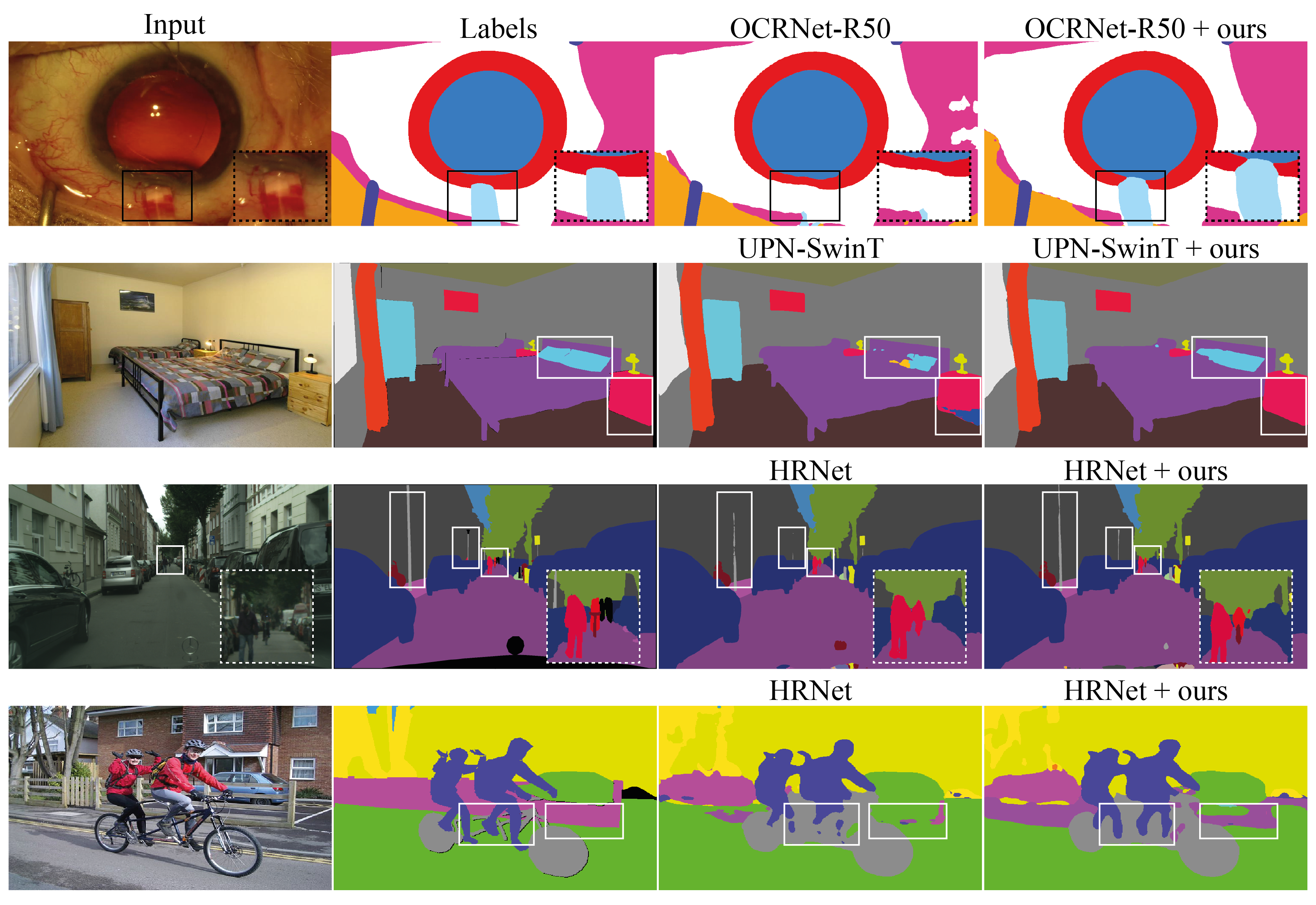}
\end{center}
\caption{\textbf{Qualitative results}: Qualitative comparisons across all $4$ datasets. More results are provided in the supplementary material.}
\label{fig:cts_qualitative}
\end{figure}
\begin{figure}[t]
\includegraphics[width=1\linewidth]{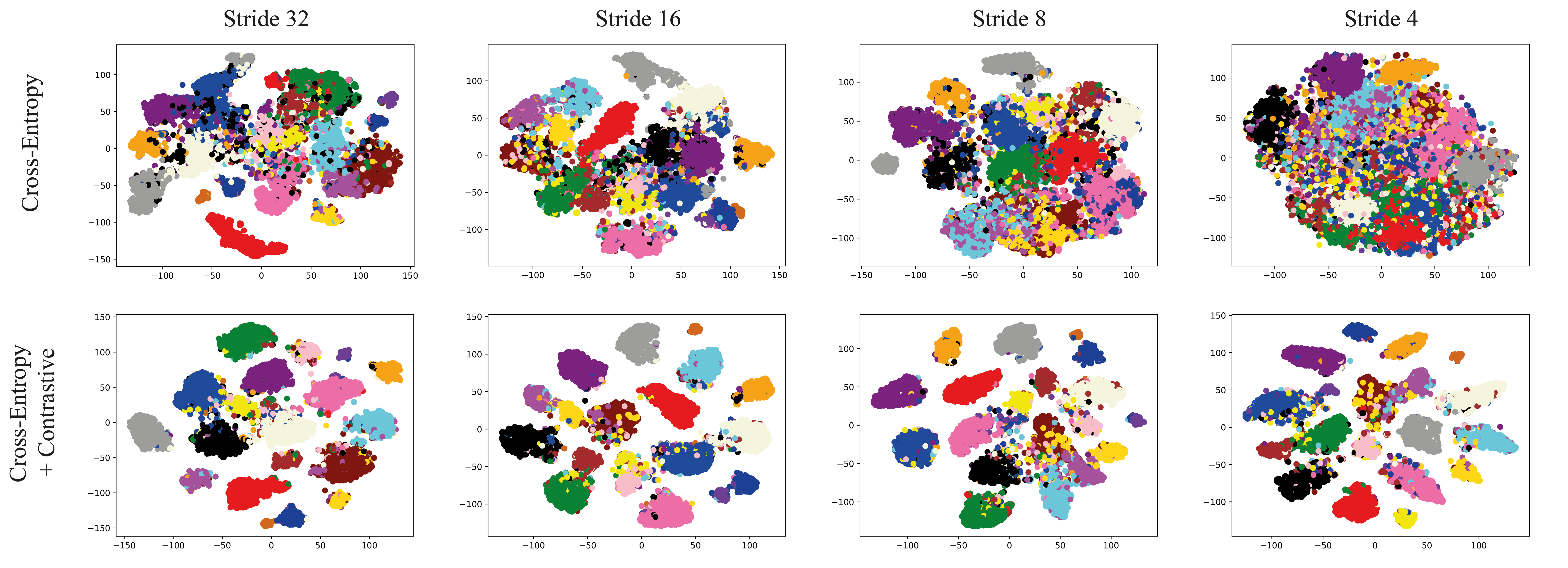}
\centering
\caption[Multi-scale and Cross-scale contrastive learning]{TSNE \cite{tsne} visualisation of the feature spaces of HRNet, on Cityscapes, trained without (top) and with (bottom) our proposed contrastive loss. Color indicates each sample's class.}
\label{fig:tsne}
\end{figure}
\subsection{Ablations and comparisons to state-of-the-art}
For ablations, we train for $40K$ steps with a batch size of $8$ and for $160K$ steps with batch size of $16$ on Cityscapes and ADE20K, respectively.

\noindent \textbf{Ablation of loss components}: First, we conduct an ablation study to demonstrate the importance of each component of our proposed loss on Cityscapes using HRNet, reported in Table \ref{tab:abl_losses_comp_aux}(a). We repeat this ablation on ADE20K using UPerNet with a Swin-T backbone, shown in Table \ref{tab:abl_losses_comp_aux}(c). We report the average mIoU across $4$ runs with different random seeds to demonstrate the stability of the ranking of methods with respect to initialization and training stochasticity.

\noindent \textbf{Comparisons to other auxiliary losses}: Tables \ref{tab:abl_losses_comp_aux}(b), \ref{tab:abl_losses_comp_aux}(d) compare the proposed loss to other auxiliary losses on Cityscapes and ADE20K, respectively.

% we utilize \textbf{Cityscapes}  as the main testbed of our method. First, we conduct an ablation study to demonstrate the importance of each component of our proposed loss using HRNet, reported in Table \ref{tab:abl_losses_comp_aux}(a) and \ref{tab:abl_losses_comp_aux}(c).
%We also examine the effect of the weights and position of our loss in Table \ref{tab:abl_settings}.

\noindent \textbf{Comparisons to contrastive learning approaches}: Table  \ref{tab:comp_other_contrastive} compares our approach to concurrent contrastive learning methods for semantic segmentation.

\noindent \textbf{Performance improvements}: In Tables \ref{tab:sota_ade20k_cadis}(a), \ref{tab:sota_ade20k_cadis}(b), \ref{tab:sota_cts} and \ref{tab:sota_pascal}, we report performance improvements obtained by training a variety of models with our loss (Eq. \ref{eq:total_loss}) (referred to as "ours") on $4$ datasets. We refer to cross-entropy as "CE" in tables. We experiment with \dv, HRNet, OCRNet and UPerNet with ResNet-101 and Swin (T,S,B,L) Transformer backbones. %References to a previously published result are accompanied by our reproduction. 
To train models using our proposed loss we leave all other settings the same as for the referenced or implemented baselines (i.e without our loss) to enable fair comparisons.

\begin{table}[t]
  \centering
    \caption{Comparison with concurrent supervised contrastive learning methods for semantic segmentation on \textbf{Cityscapes val} with single scale evaluation. Our proposed loss matches the performance of those methods without using a memory bank and relying only on anchors sampled from multiscale features.}
  \resizebox{0.95\linewidth}{!}{
    
  \begin{tabular}{*6c}
    \toprule
    \mc{Network}   & \multicolumn{3}{c}{Loss}  & \multicolumn{2}{c}{mIoU}  \\
    \cmidrule(r){1-1}                \cmidrule(){2-4}           \cmidrule(l){5-6} 
    \tbf{Model}                             & Function      & Memory            & $\ca{A}$ Selection        & @$40K$ &  Best \\
    \midrule
    \mc{\hrn} (ours)                        & \mc{$\Lc_{cms}$, $\Lc_{ccs}$}       & \mc{-}             & \mc{Multi-scale Balanced} &\mc{\textbf{81.5}} &\mc{\textbf{82.2}} \\ 
    \mc{\hrn} \cite{exploring_contrastive}  & \mc{$\Lc_{c}$}                       & \mc{Pixel}         & \mc{Hard-example}      &\mc{80.5} &\mc{-} \\ 
    \mc{\hrn} \cite{exploring_contrastive}  & \mc{$\Lc_{c}$}                       & \mc{Region+Pixel}  & \mc{Hard-example}      &\mc{81.0} &\mc{\textbf{82.2}} \\ 
    \mc{\hrn} \cite{region_aware_contrast}  & \mc{$\Lc_{c}$+aux}                   & \mc{Region}        & \mc{Averaging}         &\mc{-} &\mc{81.9} \\ 
    \bottomrule
  \end{tabular}}
  \label{tab:comp_other_contrastive}
\end{table}
 % comparison to other contrastive losse

\subsection{Implementation details}
\label{subsec:implementation}
% \todo[inline]{complete rewrite of this section}
\noindent\textbf{Contrastive Losses:} All instantiations of the loss of Eq. (\ref{eq:infoNCE}) utilize a temperature $\tau=0.1$.  A distinct projector comprising $2$ \textit{Conv}$1$x$1$\textit{-ReLU-BN} layers and a linear mapping with $d=256$, is attached to features of each scale for multi-scale and cross-scale loss variants. We utilize the $C2$, $C3$, and $C5$ features, with output strides of $4$,  $8$ and $8$, respectively, for models with a dilated ResNet backbone. When using \hrn~as the backbone, we utilize features from all $4$ scales, with output strides of $4,8,16,32$. The cross-scale loss is applied for scale pairs $(4,32)$ and $(4,16)$. When using UPerNet we attach our loss on all $4$ scales of the backbone on Cityscapes and on the feature pyramid net (FPN) outputs on ADE20K, The weights $w_{s}$ in Eq. (\ref{eq:cont_ms}) are set to $1, 0.7, 0.4, 0.1$ for feature maps of strides $4,8,16,32$ respectively in all experiments. The two latter choices are supported by ablations provided in the supplementary material.

\noindent\textbf{Training settings:} For experiments on all datasets/models, to enable fair comparisons we closely follow the training settings specified in \cite{HRNet,OCR,deeplabv3,swin} and their respective official implementations. Unless otherwise stated, on Cityscapes for CNN backbones we use SGD with a batch size of $12$ for $120$K steps while for transformer backbones we use AdamW \cite{adamw} with a batch size of $8$ for $160$K steps. The crops size used is $512\times1024$ for all models. On ADE20K and Pascal-Context, the only differences are that all models are trained with a batch size of $16$, a crop size of $512\times512$, for $160$K and $62$K steps, respectively. On CaDIS and train for $50$ epochs and use repeat factor sampling, following \cite{what_matters}. More detailed training settings are described in the supplementary material.
% On \textbf{Cityscapes}, we utilize SGD optimizer with a learning rate of $10^{-2}$ decayed with a polynomial schedule with $power=0.9$, weight decay of $10^{-4}$, images cropped to $512\times1024$ and random fliiping, color jitter and scaling with a factor in $[0.5, 2]$. For ablations we train for $108$ epochs ($\approx 40K$ steps) with a batch size of $8$. For evaluation on the val and test sets we train on $484$ epochs ($\approx 120K$ steps), On the test set we perform the standard test-time flipping and scaling during inference. On \textbf{CaDIS}, we follow the training policy of \cite{what_matters}, namely we train for $50$ epochs with a learning rate of $10^{-4}$, exponentially decayed with a rate $0.96$, use the Lovasz loss \cite{Lovasz} combined with our contrastive loss and repeat factor sampling. On \textbf{Pascal-Context} we use a batch size of $16$ and a polynomially decayed learning rate, weight decay of $10^{-4}$.   

\section{Results and Discussion}
\label{sec:results}

As shown in Tables \ref{tab:abl_losses_comp_aux}(a) and \ref{tab:abl_losses_comp_aux}(c), using both multi- and cross-scale contrastive terms provides the highest improvement relative to the baseline. Additionally, as shown in Tables \ref{tab:abl_losses_comp_aux}(b) and \ref{tab:abl_losses_comp_aux}(d), our loss outperforms other auxiliary losses.

\begin{table}[ht]
\centering
\captionsetup[subfloat]{position=top}
  \caption{(a): Comparisons on \textbf{ADE20K} with single/multi-scale evaluation (ss/ms). If no reference is provided the value is obtained by our implementation and OHEM refers to the loss of \cite{ohem}. (b): Comparisons on \textbf{CaDIS} tasks $2$ and $3$ that define $17$ and $25$ classes respectively. $\dagger$ : Imagenet-$22$K pretraining.} 
  \label{tab:sota_ade20k_cadis}
\subfloat[Subtable 1 list of tables text][ADE20K]{
        \resizebox{0.7\linewidth}{!}{
  \begin{tabular}{*7c} % c|c|c|c|c|c|c
    \toprule
    \multicolumn{4}{c}{Network}    & \mc{Loss}                   & \multicolumn{2}{c}{mIoU}  \\ 
    
    \cmidrule(r){1-4}               \cmidrule(){5-5}              \cmidrule(l){6-7}  
    
    \tbf{Model}      & \tbf{Backbone}  &\tbf{$\#$Params }(M)   &\tbf{Source}          & \quad            & ss/ms       & Improvement \\
    \midrule
    \mc{\dv}      & \mc{R$101$} & 63 & \cite{resnetst}                      & \mc{CE}                      & \mc{\qquad-/44.1}   & \quad    \\
    \mc{\dv}      & \mc{R$101$} & 63 & -                                    & \mc{ours}                    & \mc{\textbf{44.2/45.6}}      & $\color{green}(-/+1.5)$    \\  
    \mc{\dv}      & \mc{R$101$} & 63 & \cite{region_aware_contrast}         & \mc{$\Lc_{c}$+Mem+OHEM}      & \mc{\qquad-/46.8}   & \quad    \\

    \midrule

    \mc{OCRNet}      & \mc{HR$48$v$2$} & 71 & \cite{OCR}         & \mc{OHEM}        & \mc{44.5/45.5}                   & \quad    \\ 
    \mc{OCRNet}      & \mc{HR$48$v$2$} & 71 & -                  & \mc{ours}        & \mc{\textbf{46.8}/\textbf{47.4}} & $\color{green}(+2.3/+1.9)$ \\ 

    \midrule
    \mc{UPerNet}      & \mc{R$101$} & 86   & \cite{UPerNet}      & \mc{CE}          & \mc{42.0/42.7}                & \quad    \\ 
    \mc{UPerNet}      & \mc{R$101$} & 86   & -                   & \mc{ours}        & \mc{\tbf{43.8}/\textbf{45.3}} & $\color{green}(+1.8/+2.6)$\\ 

    \midrule

    \mc{UPerNet}      & \mc{Swin-T} & 60   & -                   & \mc{CE}          & \mc{44.7/45.5}       & \quad    \\ 
    \mc{UPerNet}      & \mc{Swin-T} & 60   & \cite{swin}         & \mc{CE}          & \mc{44.5/45.8}       & \quad    \\ 
    \mc{UPerNet}      & \mc{Swin-T} & 60   & -                   & \mc{ours}        & \mc{\tbf{45.9}/\textbf{46.6}}             & $\color{green}(+1.4/+0.8)$ \\ 
    \midrule

    \mc{UPerNet}      & \mc{Swin-S} & 81 & -                   & \mc{CE}          & \mc{48.1/49.2}     & \quad    \\ 
    \mc{UPerNet}      & \mc{Swin-S} & 81 & \cite{swin}         & \mc{CE}          & \mc{47.6/49.5}     & \quad    \\ 
    \mc{UPerNet}      & \mc{Swin-S} & 81 & -                   & \mc{ours}        & \mc{\textbf{48.9}/\textbf{50.0}}     & $\color{green}(+1.3/+0.5)$ \\ 

    \midrule
    \mc{UPerNet}      & \mc{Swin-B$\dagger$} & 121 & \cite{swin}         & \mc{CE}          & \mc{50.1/51.6}  & \quad    \\ 
    \mc{UPerNet}      & \mc{Swin-B$\dagger$} & 121 & -                   & \mc{ours}        & \mc{\textbf{51.3}/\textbf{52.2}}       & $\color{green}(+1.2/+0.6)$ \\ 
    % dgx 20220304_020537_e1__upn_alignFalse_projFpn_22k_swinB_sbn_DCms_cs_epochs127_bs16
    \midrule
    \mc{UPerNet}      & \mc{Swin-L$\dagger$} & 234 & \cite{swin}         & \mc{CE}        & \mc{52.0/53.5}       & \quad \\ 
    \mc{UPerNet}      & \mc{Swin-L$\dagger$} & 234 & -        & \mc{ours}          & \mc{\textbf{52.9}/53.3}       & $\color{green}(+0.9/\color{red}-0.2)$  \\ 

    \bottomrule
  \end{tabular}
      } % resize box
    }
\qquad
\subfloat[CaDIS][CaDIS]{
        \resizebox{0.7\linewidth}{!}
        { %< auto-adjusts font size to fill line
         \begin{tabular}{@{}lccccccccccc@{}}
                \toprule
                \multicolumn{3}{c}{Method} & \multicolumn{2}{c}{mIoU} & \multicolumn{2}{c}{Anatomies} & \multicolumn{2}{c}{Tools} & \multicolumn{2}{c}{Rare}\\ 
                \cmidrule(r){1-3}            \cmidrule(l){4-5}          \cmidrule(l){6-7}              \cmidrule(l){8-9}            \cmidrule(l){10-11}
                \tbf{Model} & \tbf{Backbone} & \mc{Loss}  & Task2          & Task3             & Task2              & Task3          & Task2          & Task3          & Task2          & Task3 \\
                \midrule
                OCRNet         & R50            & CE         & 81.02          & 76.24             & \textbf{90.80}     & 90.87          & 74.65          & 70.82          & 74.46          & 67.58\\
                OCRNet         & R50            & +ours      & \textbf{81.47} & \textbf{77.67}    & 90.66              & \textbf{90.91} & \textbf{75.58} & \textbf{72.79} & \textbf{77.63} & \textbf{73.09}\\
                Improvement     & \quad              & \quad     & $\color{green}(+0.45)$ & $\color{green}(+1.43)$ & $\color{red}(-0.14)$ & $\color{green}(+0.04)$ & $\color{green}(+0.97)$ & $\color{green}(+1.97)$ & $\color{green}(+3.17)$ & $\color{green}(+5.51)$\\
                \midrule
                OCRNet & R50                      & Lovasz    & 82.36 & 77.77 & \textbf{90.63} & 90.59 & 76.89 & 72.96 & 77.52 & 71.44\\
                OCRNet & R50                      & +ours      & \textbf{82.56} & \textbf{78.25} & 90.59 & \textbf{90.76} & \textbf{77.41} & \textbf{73.11} & \textbf{78.55} & \textbf{72.67}\\
                \midrule
                Improvement & \quad            & \quad      & $\color{green}(+0.20)$ & $\color{green}(+0.48)$ & $\color{red}(-0.04)$ & $\color{green}(+0.17)$ & $\color{green}(+0.52)$ & $\color{green}(+0.15)$ & $\color{green}(+1.03)$ & $\color{green}(+1.23)$\\
                \bottomrule
            \end{tabular}
        } % resize box
  }
\end{table}

\begin{table}[ht]
\centering
\captionsetup[subfloat]{position=top}
  \caption{(a): Comparisons with strong baselines on \textbf{Cityscapes val} with single scale evaluation (ss). If no reference is provided the value is obtained by our implementation. The denoted improvements are relative to the baseline with the highest mIoU between our implementation and previously reported results.  (b): Comparison on \textbf{Cityscapes test} with all models trained on \textbf{train}+\tbf{val}.} 
\subfloat[Subtable 1 list of tables text][Cityscapes val]{
        \resizebox{0.45\linewidth}{!}{
        \begin{tabular}{*6c}
        \toprule
        \multicolumn{3}{c}{Network}    & \mc{Loss}                   & \multicolumn{2}{c}{mIoU} (ss)  \\ 
        
        \cmidrule(r){1-3}               \cmidrule(l){4-4}              \cmidrule(l){5-6}  
        
        \tbf{Model}      & \tbf{Backbone}  &\tbf{Source}          & \quad            & \quad           & Improvement                \\
        \midrule    
        \mc{PSPNet}     & \mc{R$101$} & \cite{segsort2019}      & \mc{Metric-learning}          & \mc{78.2}                                      \\ 
        \midrule    
        \dv             & \mc{R$101$} & \cite{deeplabv3}        & \mc{CE}          & \mc{77.8}                                      \\ 
        \dv             & \mc{R$101$} & -                       & \mc{CE}          & \mc{78.2}                                      \\ 
        \dv             & \mc{R$101$} & -                       & \mc{ours}        & \mc{\tbf{79.0}} & $\color{green}(+0.8)$        \\
        
        \midrule    
    
        \mc{\hrn}        & \mc{HR$48$v$2$} & -                  & \mc{CE}          & \mc{81.0}       & \quad                  \\
        \mc{\hrn}        & \mc{HR$48$v$2$} & \cite{HRNet}       & \mc{CE}          & \mc{81.1}       & \quad             \\
        \mc{\hrn}        & \mc{HR$48$v$2$} & -                  & \mc{ours}        & \mc{\tbf{82.2}} & $\color{green}(+1.1)$  \\
        \mc{\hrn}   & \mc{HR$48$v$2$} & \cite{exploring_contrastive} & \mc{$\Lc_{c}$+Mem}  & \mc{\tbf{82.2}} & \quad       \\
       
        \midrule    
    
        \mc{OCR}        & \mc{HR$48$v$2$} & -                  & \mc{CE}          & \mc{81.2}       & \quad                    \\
        % \mc{OCR}        & \mc{HR$48$v$2$} & \cite{mmseg}       & \mc{CE}          & \mc{81.3}       & \quad                    \\
        \mc{OCR}        & \mc{HR$48$v$2$} & \cite{OCR}         & \mc{CE}          & \mc{81.6}       & \quad                    \\
        \mc{OCR}        & \mc{HR$48$v$2$} & -                  & \mc{ours}        & \mc{\tbf{81.9}} & $\color{green}(+0.3)$    \\

        \midrule    
        
        \mc{UPerNet}     & \mc{R$101$}     & -                  & \mc{CE}          & \mc{78.0}                  & \quad                    \\
        \mc{UPerNet}     & \mc{R$101$}     & -                  & \mc{ours}        & \mc{\textbf{79.1}}         & $\color{green}(+1.1)$    \\
        
        \midrule
        
        \mc{UPerNet}     & \mc{Swin-T}     & -                  & \mc{CE}          & \mc{79.2}                  & \quad                    \\
        \mc{UPerNet}     & \mc{Swin-T}     & -                  & \mc{ours}        & \mc{\textbf{79.9}}         & $\color{green}(+0.7)$    \\
        
        \midrule
    
        \mc{UPerNet}     & \mc{Swin-S}     & -                  & \mc{CE}          & \mc{80.9}                  & \quad                    \\
        \mc{UPerNet}     & \mc{Swin-S}     & -                  & \mc{ours}        & \mc{\textbf{81.7}}         & $\color{green}(+0.8)$    \\
        \midrule

        \mc{UPerNet}     & \mc{Swin-B}     & -                  & \mc{CE}          & \mc{82.0}                  & \quad                    \\
        \mc{UPerNet}     & \mc{Swin-B}     & -                  & \mc{ours}        & \mc{\textbf{82.6}}         & $\color{green}(+0.6)$    \\
        \bottomrule
      \end{tabular}
      } % resize box
    }
\qquad
\subfloat[Subtable 2 list of tables text][Cityscapes test]{
        \resizebox{0.45\linewidth}{!}
        { %< auto-adjusts font size to fill line
        \begin{tabular}{*6c}
        \toprule
        Method & Loss & mIoU & iIoU & IoU Cat & iIoU Cat \\        \midrule
        \hrn \cite{HRNet} &  CE & 81.6 & 61.8 & 92.1 & 82.2 \\
        \hrn & ours        & \tbf{81.9} & \tbf{62.9} & \tbf{92.2} & \textbf{83.4}  \\
        Improvement        & \quad & $\color{green}(+0.3)$ & $\color{green}(+1.1)$ & $\color{green}(+0.1)$  & $\color{green}(+1.2)$  \\
        % run_id : 20211225_225551_e1__hrn_feats20K_hr48_sbn_DCms_cs
        % link to best result https://www.cityscapes-dataset.com/anonymous-results/?id=69173d335cd66516e92825731201fa9b6755d007e192d169271e17c977ef397c
        
        \midrule
        OCRNet \cite{OCR} & CE   & \textbf{82.5} & 61.7 & 92.1 & 81.6  \\
        OCRNet            & ours & 82.4 &\textbf{63.6} & \textbf{92.2}  &\textbf{83.7}  \\
        Improvement           & \quad & $\color{red}(-0.1)$ & $\color{green}(+1.9)$ & $\color{green}(+0.1)$  & $\color{green}(+2.1)$  \\
        % run_id : 20220103_003614_e1__hrnOCR_hr48_sbn_DCms_cs_epochs415_bs12
        % https://www.cityscapes-dataset.com/submit/?action=anonymousLink&submissionID=13979
        
        \bottomrule
        \end{tabular}
        } % resize box
  }
\vspace{5mm}

  \label{tab:sota_cts}
\end{table}
  
\begin{table*}[htb]
  \centering
  \caption{Comparisons on \textbf{Pascal-context val} with multi-scale evaluation (ms).}
\resizebox{0.5\linewidth}{!}{
  \begin{tabular}{*6c}
    \toprule
    \multicolumn{3}{c}{Network}    & \mc{Loss}                   & \multicolumn{2}{c}{mIoU (ms)}  \\ 
    
    \cmidrule(r){1-3}               \cmidrule(){4-4}              \cmidrule(l){5-6}  
    
    \tbf{Model}      & \tbf{Backbone}  &\tbf{Source}          & \quad            & \quad           & Improvement \\
    
    \midrule    

    % \hrn             & \mc{HR$48$v$2$} & \cite{mmseg}       & \mc{CE}          & \mc{53.5}       & \quad    \\ 
    \hrn             & \mc{HR$48$v$2$} & -                  & \mc{CE}          & \mc{53.5}       & \quad    \\ 
    \hrn             & \mc{HR$48$v$2$} & \cite{HRNet}       & \mc{CE}          & \mc{54.0}       & \quad    \\ 
    \hrn             & \mc{HR$48$v$2$} & -                  & \mc{ours}        & \mc{\textbf{55.3}}       & $\color{green}(+1.3)$ \\
    \midrule
    \hrn             & \mc{HR$48$v$2$} & \cite{exploring_contrastive} & \mc{$\Lc_{c}$+Mem}        & \mc{55.1}       & \quad    \\
    \midrule    
    \mc{OCRNet}      & \mc{HR$48$v$2$} & -                  & \mc{CE}          & \mc{55.8}       & \quad    \\ 
    \mc{OCRNet}      & \mc{HR$48$v$2$} & \cite{OCR}         & \mc{CE}          & \mc{56.2}       & \quad    \\ 
    \mc{OCRNet}      & \mc{HR$48$v$2$} & -                  & \mc{ours}        & \mc{56.5}       & $\color{green}(+0.3)$ \\ 
    \midrule
    \mc{OCRNet}      & \mc{HR$48$v$2$} & \cite{exploring_contrastive} & \mc{$\Lc_{c}$+Mem}       & \mc{\textbf{57.2}}       & \quad    \\

    \bottomrule
  \end{tabular}
  
  } % resizebox  
  \label{tab:sota_pascal} 
\end{table*}

Regarding \textbf{other supervised contrastive losses}, on \textbf{Cityscapes}, where we are able to provide a comparison between all methods, our loss is competitive, matching \cite{exploring_contrastive} and outperforming \cite{region_aware_contrast}, despite not using a memory bank, hard-example mining or region averaging (Table \ref{tab:comp_other_contrastive}). On \textbf{ADE20K}, \cite{region_aware_contrast} only report results with \dv~ and achieve higher improvement over the baseline than ours ($\mathbf{+2.7\%}$ vs $\mathbf{+1.5\%}$), albeit using both memory, OHEM and an intermediate auxiliary loss. Importantly, we showcase improvements for a wider range of models on this dataset. On \textbf{Pascal-Context}, we marginally outperform \cite{exploring_contrastive} when using \hrn\ and are outperformed when using OCRNet. 

Notably, our loss improves performance for various CNN and Transformer-based models across datasets: On the challenging \textbf{ADE20K} dataset, (Table \ref{tab:sota_ade20k_cadis}(a)), we obtain an improvement for OCRNet by $\mathbf{+2.3\%}$ (ss) and $\mathbf{+1.9\%}$ (ms), for UPerNet by $\mathbf{+1.8\%}$ (ss) and $\mathbf{+2.6\%}$ (ms) and for \dv\ by $\mathbf{+1.5\%}$ (ms). We also improve the state-of-the-art model of \cite{swin} using UPerNet with Swin Transformer for increasing backbone sizes (Swin-T,-S,-B, -L) by $\mathbf{+1.4\%}$, $\mathbf{+1.3\%}$, $\mathbf{+1.2\%}$ and $\mathbf{+0.9\%}$ (ss). Overall, with only a single-scale input, CNN models and Swin variants trained with our loss achieve performance close to that of the baseline when the latter employs the computationally expensive (several seconds per image) multiscale test-time augmentation.

% Notably, the (ms) performance when using Swin-B outperforms the result of \cite{swin} achieving $\mathbf{52.2}$, thus closing the gap to Swin-L from $1.9$ to $1.3$ despite using almost half the parameters ($121$M vs $234$M). 

On \textbf{Cityscapes-val} (Table \ref{tab:sota_cts}), we improve both \hrn\ and UPerNet with ResNet$101$ by $\mathbf{+1.1\%}$, \dv\ by $\mathbf{+0.8\%}$ and UperNet with Swin-T and Swin-S by $\mathbf{+0.7\%}$ and $\mathbf{+0.8\%}$ respectively. Our approach with \hrn\ and OCRNet, did not significantly improve mIoU on the \textbf{test} set but in both cases outperforms baselines reported in \cite{HRNet,OCR} in terms of \textbf{iIoU} and \textbf{iIoU-Cat} (Table \ref{tab:sota_cts}(b)) by notable margins. This can be attributed to our sampling approach, that balances loss contributions from all present classes (Sec.\ \ref{subsec:sampling}) and the fact that our loss enhances the discriminative power of features at multiple scales, as exemplified in Figure\ \ref{fig:tsne}.
 
On \textbf{Pascal-Context}, adding our loss to \hrn\ leads to $\mathbf{+1.3\%}$ and a small improvement for OCRNet (Table \ref{tab:sota_pascal}) while doing so on the method of \cite{what_matters}, on \textbf{CaDIS} results in state-of-the-art performance, especially favouring the rarest of classes, respectively for tasks $2$ and $3$, by $\mathbf{+1.0\%}$ and $\mathbf{+1.2\%}$, when combined with the Lovasz loss, and by $\mathbf{+3.1\%}$ and $\mathbf{+5.2\%}$ when combined with CE. While the mean mIoU is a standard metric for assessing semantic segmentation, focusing it over the rarest classes is crucial to assess long-tailed class performance. This is especially important in the surgical domain where collecting data of rarely appearing tools, under real surgery conditions, can be particularly difficult. % while their recognition is aggravated by many surgical tools being almost identical in appearance.

\section{Conclusion}
% \todo{also write about limitations + societal impact}

%\noindent \textbf{Conclusion}: 
We presented an effective method for supervised contrastive learning both at multiple feature scales and across them. %To this end, we propose a novel contrastive loss term that endows the model with cross-scale semantic links between its features. 
Overall, we showcased significant gains for most of the strong CNN and Transformer-based models we experimented with on $4$ datasets. Notably, our approach achieved maximal gains on the challenging ADE20K dataset, which contains a large number semantic concepts ($150$ classes), where the recognition component of segmentation greatly benefits from the class-aware clustered feature spaces of our method. %This can be ascribed to the intuition that with a high number of classes, recognition becomes more challenging, and our losses contribute in obtaining more tightly clustered feature spaces and increasingly discriminative features across multiple layers of the model. %When considering datasets with less classes where we especially observed improved performance around small-sized (Cityscapes) object classes or classes with very low frequency in the dataset (CaDIS), hinting our .

%One limitation of our method is that 

% One limitation of our method is that the utilized contrastive loss terms require pixel-wise labels. Additionally, we also leave for future work the exploration of our method on transformer architectures, which recently emerged as a very effective alternative for dense prediction tasks, but remain computationally expensive. %\todo{Something on the baseline}.

\noindent \textbf{Acknowledgements}: This work was supported by an ERC Starting Grant [714562].% and by core funding from the Wellcome/EPSRC Centre for Medical Engineering, Wellcome Trust [WT203148/Z/16/Z].

\newpage
\appendix

% --- PDF will be split by an editor (e.g. macOS preview), so need to restart from page 1
% \setcounter{page}{1}

% --- repeat the title (AT: haven't found a more elegant way to do this...)
% \twocolumn[
% \centering

\begin{comment}
\begin{center}
\Large

\textbf{Multi-scale and Cross-scale Contrastive Learning for Semantic Segmentation
} \\
\vspace{0.5em}Supplementary Material \\
\vspace{1.0em}
\end{center}
% ] %< twocolumn
\end{comment}

\begin{center}
\Large
\textbf{Supplementary Material} \\
% \vspace{0.5em} \\
% \vspace{1.0em}
\end{center}
% ] %< twocolumn

\appendix
\small

% \section{Potential Negative Impact}

% Our work proposes a general method for improving performance of semantic segmentation systems. Such advancements can be beneficial for applications such as driving autonomy through better urban scene understanding, or in clinical settings by providing assistance to surgeons by identifying targeted anatomies and employed tools. However, it is possible that better semantic segmentation are utilized for tracking in military settings. Additionally, an important, potentially harmful, limitation of our model and similar models in the literature, is that their inferences regarding the input scene are biased to the dataset and the loss used for training them, the latter leveraging the maximum-likelihood principle. For example, training on CITYSCAPES biases the model by only considering western European urban environments, leaving it ineffective and dangerous for deployment in other urban settings across the world, with varying cultural and environmental characteristics. Finally, while we show improved performance on a surgical dataset, our model or any similarly trained/evaluated model is far from ready for clinical use as it has not been tested on data from different surgical sites and different patient populations to ensure safety.

\section{Selecting loss hyperparameters}
The proposed loss (Eq. (5)) requires selecting certain hypeparameters, namely: the number of feature scales, the choice of cross-scale pairs, the per scale and overall weights of contrastive losses. Our results are obtained with minimal model or dataset-specific tuning of those parameters. Specifically, for \textbf{all models and datasets} we set both weights of Eq.(5) to $0.1$ and use $2$ scale-pairs (s4-s32, s4-s16) based on the results of the ablation in Tables 1(a) and 1(c) of the main paper. We further tested two different per-scale weight and cross-scale pair choices using a single model (HRNet) on Cityscapes (Table 1.b) and adopt per-scale weights as a decreasing function of the output stride. Finally, we tested two different alternatives regarding the position of the loss when using the UPerNet architecture, where the loss can be applied either on the FPN outputs or the directly over the backbones features. For Cityscapes the optimal choice is the latter while on ADE20K it is the former (Table 1(a)) and we adopt these choices for all other experiements on each dataset when using UPerNet. Thus, with minimal tuning our approach is effective while potentially further model- or dataset-specific tuning can boost performance even more.

\begin{table}[ht]
\centering
\caption{Ablation on (a) the position of application of the multi-scale and cross-scale losses for models using the UPerNet architecture and (b) on values of weights $w_s$ of the multi-scale loss of  Eq.~(4).}\label{tab:abl_settings}
\subfloat[Subtable 2 list of tables text][]{
\resizebox{68mm}{!}{
  \begin{tabular}{*5c}
    \toprule
    \multicolumn{2}{c}{Network}                                 & \mc{Loss position}  & \mc{Dataset}     &\mc{mIoU} (ss)  \\
    \cmidrule(r){1-2}                                           \cmidrule(l){3-3}   \cmidrule(l){4-4}  \cmidrule(l){5-5}
    % \tbf{Model}       & \tbf{Backbone}                          & \quad             & \quad                  \\
    
    \mc{UPerNet}      & \mc{R101}                               & Backbone          & CTS      &\mc{\tbf{79.1}}  \\ 
    \mc{UPerNet}      & \mc{R101}                               & FPN               & CTS      &\mc{78.4}        \\ 
    
    \midrule
    
    \mc{UPerNet}      & \mc{Swin-S}                             & Backbone          & CTS      &\mc{\tbf{81.7}}  \\ 
    \mc{UPerNet}      & \mc{Swin-S}                             & FPN               & CTS      &\mc{80.9}  \\ 
    
    \midrule
    \mc{UPerNet}      & \mc{Swin-S}                             & Backbone          & ADE20K          &\mc{47.9}  \\    
    \mc{UPerNet}      & \mc{Swin-S}                             & FPN          & ADE20K          &\mc{\textbf{49.0}}  \\

    \bottomrule
  \end{tabular}}}
\qquad
\subfloat[Subtable 1 list of tables text][]
  {
    \resizebox{36mm}{!}{
      \begin{tabular}{c|c}
        \toprule
        \mc{$w_s$}                                  & \mc{mIoU (ss)}     \\
        \midrule
         \mc{$1.0$ $1.0$ $1.0$ $1.0$}            & \mc{81.8}       \\ 
         \mc{$1.0$  $0.7$ $0.4$ $0.1$}           & \mc{\tbf{82.2}} \\
        \bottomrule
      \end{tabular}
      }
  }
\qquad
 %\textbf{Cityscapes val}
\end{table}

\section{Additional ablations}

We report additional ablations regarding the effect of using longer training schedules and the importance of using the sampling process described in Section $3.3$. As can be seen our method benefits by a longer training schedule and a bigger batch size while staying ahead of the baseline in all cases (Table \ref{tab:settings_supp}(b)). Further, as shown in Table \ref{tab:settings_supp}(a), our use of anchor sampling is necessary to allow an extension of contrastive losses to multiple scales as memory consumption exceeds our utilized hardware's capacity ($4\times24$GB-GPUs). Further we find that even with a single contrastive loss term ($\Lc_c$) our choice to perform anchor sampling results in better performance than using all available anchors in the batch which is equivalent to obtaining a number of anchors per class (denoted by $\textbf{K}$) according to the class distribution $p_{data}$, which is imbalanced.

\begin{table}[ht]
\centering
\caption{(a) Comparison with alternative sampling options ($40$K steps, with a batch size of $8$ and using $4$). We denote the number of samples per class by $\textbf{K}$. (b) Ablation of training schedules/batch sizes. All results are on \textbf{Cityscapes val} using single scale evaluation.}\label{tab:settings_supp}
\subfloat[Subtable 2 list of tables text][]{
  \resizebox{0.99\linewidth}{!}{ %< auto-adjusts font size to fill line
\begin{tabular}{@{}lcccccccc@{}}
\toprule 
Model             & Scales  & Scale Pairs  & Loss      & Sampling    & \textbf{K} & mIoU & Mem/GPU (GB)\\
\midrule
\hrn              & 1    & -           & $\Lc_{c}$ & \quad        & $\thicksim p_{data}$                       & 79.4  & 14.2  \\
\hrn              & 1    & -           & $\Lc_{c}$ & \mc{\tick}   & Sec. 3.3                 & 80.2  & 7.4\\
\midrule
\hrn              & 4    & 2           & $\Lc_{cms}+\Lc_{ccs}$ & \quad        & $\thicksim p_{data}$           & -     & OOM\\
\hrn              & 4    & 2           & $\Lc_{cms}+\Lc_{ccs}$ & \mc{\tick}   & Sec. 3.3     & 81.5  & 9.7\\
\bottomrule
\end{tabular}
}
}
\qquad
\subfloat[Subtable 2 list of tables text][]{
\resizebox{0.5\linewidth}{!}{
    \begin{tabular}{*6c}
    \toprule
    \multicolumn{2}{c}{Network}   &  \mc{Settings}              &       \multicolumn{3}{c}{mIoU}  \\
    \cmidrule(r){1-2}                \cmidrule(l){3-3}                  \cmidrule(l){4-6} 
    \mc{Model} & \mc{Loss}              & Batch                         & \mc{40K}           & \mc{80K}           & \mc{120K}  \\
    \midrule
    \mc{\hrn}  & CE                     & \mc{8}                        & \mc{79.1}          & \mc{79.7}          & \mc{80.5} \\
    \mc{\hrn}  & +ours                  & \mc{8}                        & \mc{\textbf{81.5}} & \mc{\textbf{81.7}} & \mc{\textbf{81.6}} \\
    \midrule
    \mc{\hrn}  & CE                     & \mc{12}                       & -              & -              & \mc{81.0} \\
    \mc{\hrn}  & +ours                  & \mc{12}                       & -              & -              & \mc{\textbf{82.2}} \\
    \bottomrule
    \end{tabular}
    }
    } % resizebox
\end{table}

% \goodbreak

\newpage

\section{Additional comparisons on ADE20K}
We provide more comparisons between our results using UPerNet with Swin backbones and other state-of-the-art transformer models, on ADE20K. Notably, our result using Swin-B outperforms other competitive models despite having close to a third of the parameters in comparison to Segmenter \cite{segmenter} and SETR \cite{setr}.

\begin{table*}[htb]
  \centering
  \caption{Additional results and comparisons with SOTA on \textbf{ADE20K val}.}
        \resizebox{0.99\linewidth}{!}{
  \begin{tabular}{*7c} % c|c|c|c|c|c|c
    \toprule
    \multicolumn{4}{c}{Network}    & \mc{Loss}                   & \multicolumn{2}{c}{mIoU}  \\ 
    
    \cmidrule(r){1-4}               \cmidrule(){5-5}              \cmidrule(l){6-7}  
    
    \tbf{Model}      & \tbf{Backbone}  &\tbf{$\#$Params }(M)   &\tbf{Source}          & \quad            & ss/ms       & Improvement \\

    \mc{UPerNet}      & \mc{Swin-B$\dagger$} & 121 & \cite{swin}         & \mc{CE}          & \mc{50.1/51.6}  & \quad    \\ 
    \mc{UPerNet}      & \mc{Swin-B$\dagger$} & 121 & -                   & \mc{ours}        & \mc{\textbf{51.3}/\textbf{52.2}}  & $\color{green}(+1.2/+0.6)$  \\ 
    
    \midrule
    \mc{UPerNet}      & \mc{Swin-L$\dagger$} & 234 & \cite{swin}         & \mc{CE}        & \mc{52.0/53.5}       & \quad \\ 
    \mc{UPerNet}      & \mc{Swin-L$\dagger$} & 234 & -        & \mc{ours}          & \mc{\textbf{52.9}/53.3}       & $\color{green}(+0.9/\color{red}-0.2)$  \\ 
    
    % dgx 20220304_020537_e1__upn_alignFalse_projFpn_22k_swinB_sbn_DCms_cs_epochs127_bs16 $\color{green}(+1.2/+0.6)$ 
    \midrule
    \mc{SegFormer}    & \mc{MiT-B5} & 84   & \cite{segformer}          & \mc{CE}          & \mc{51.1/51.8}  & \quad    \\
    % \mc{UPerNet}      & \mc{Swin-L$\dagger$} & 234 & \cite{swin}         & \mc{CE}          & \mc{52.0/53.5}       & \quad \\ 
    \mc{Segmenter}    & \mc{ViT-L/16$\dagger$} & 307 & \cite{segmenter}    & \mc{CE}          & \mc{50.7/52.2}       & \quad \\ 
    \mc{SETR}    & \mc{T-Large$\dagger$} & 310 & \cite{setr}            & \mc{CE}          & \mc{48.6/50.3}       & \quad \\ 

    \bottomrule
  \end{tabular}
      } % resize box

  \label{tab:ade20k_supp} 
\end{table*}

% \newpage

\begin{figure}[t]
% \begin{center}
% \begin{overpic} 
% [width=\linewidth]
% {example-image-a}
% \end{overpic}
% \includegraphics[width=\linewidth]{example-image-golden}
% \includegraphics[width=\linewidth]{fig/fig1.png} % test 
\includegraphics[width=1\linewidth]{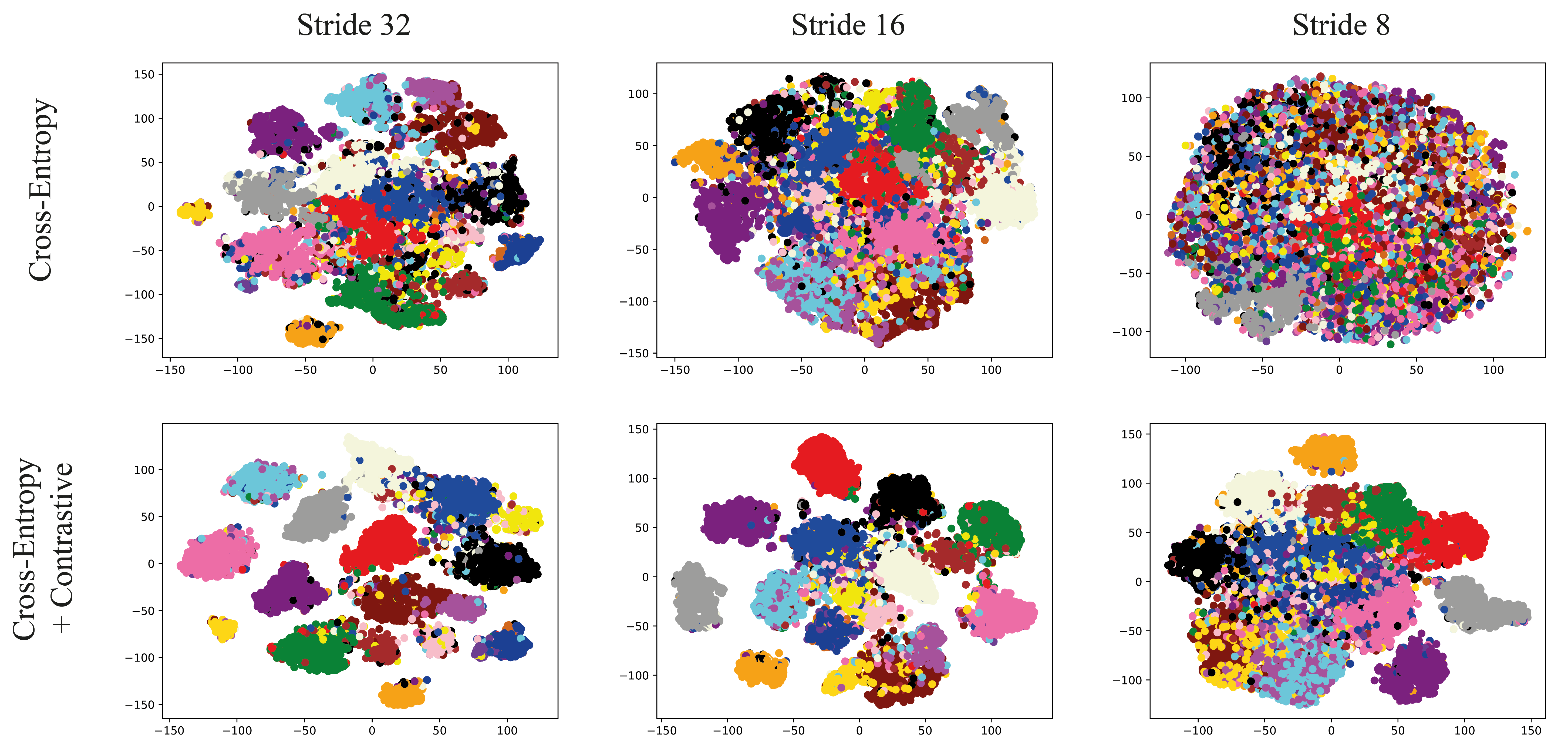}
\centering
% \end{center}
\caption[Multi-scale and Cross-scale contrastive learning]{TSNE \cite{tsne} visualisation of the feature spaces of UPerNet with ResNet-$101$ backbone, on Cityscapes, trained without (top) and with (bottom) our proposed contrastive loss. Color indicates each sample's class.}
\label{fig:tsne_supp}
\end{figure}

% % \begin{center}
% % \includegraphics[width=.99\linewidth]{fig/fig5a_supp-01-01-01.eps}
% \includegraphics[width=.99\linewidth]{fig/fig5a_supp-01-01-01-01.png}

% % \includegraphics[width=.99\linewidth]{fig/fig5a_supp-
% % \vfill
% % \includegraphics[width=.99\linewidth]{fig/fig5b_supp-02-02-02-02.eps}
% \includegraphics[width=.92\linewidth]{fig/fig5b_supp-02-02-02-02.png}
% \end{center}

\section{Qualitative results}
\begin{figure*}
\begin{center}
\includegraphics[width=.99\linewidth]{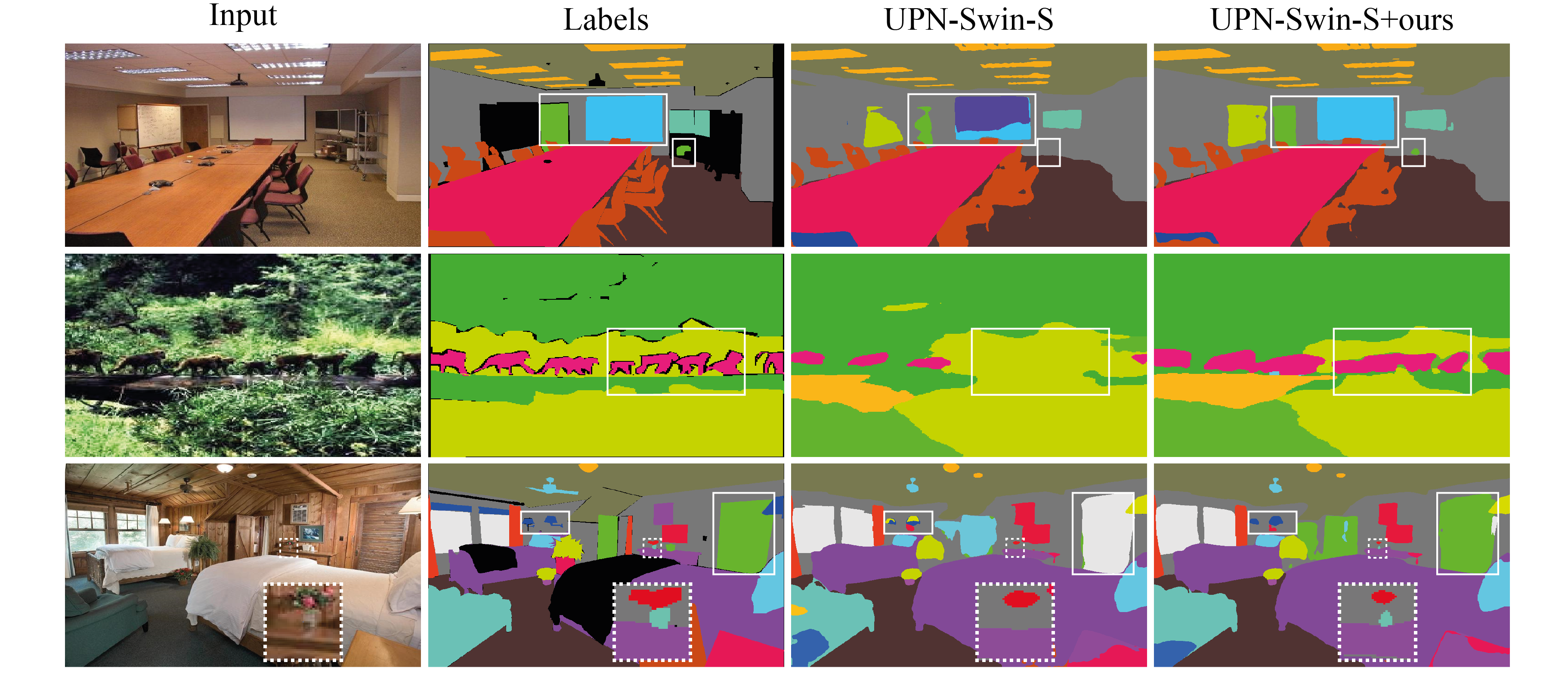}
\vfill
% \vspace{-1cm}
% \includegraphics[width=.99\linewidth]{fig/fig7-ade20k-a.eps}
\includegraphics[width=.99\linewidth]{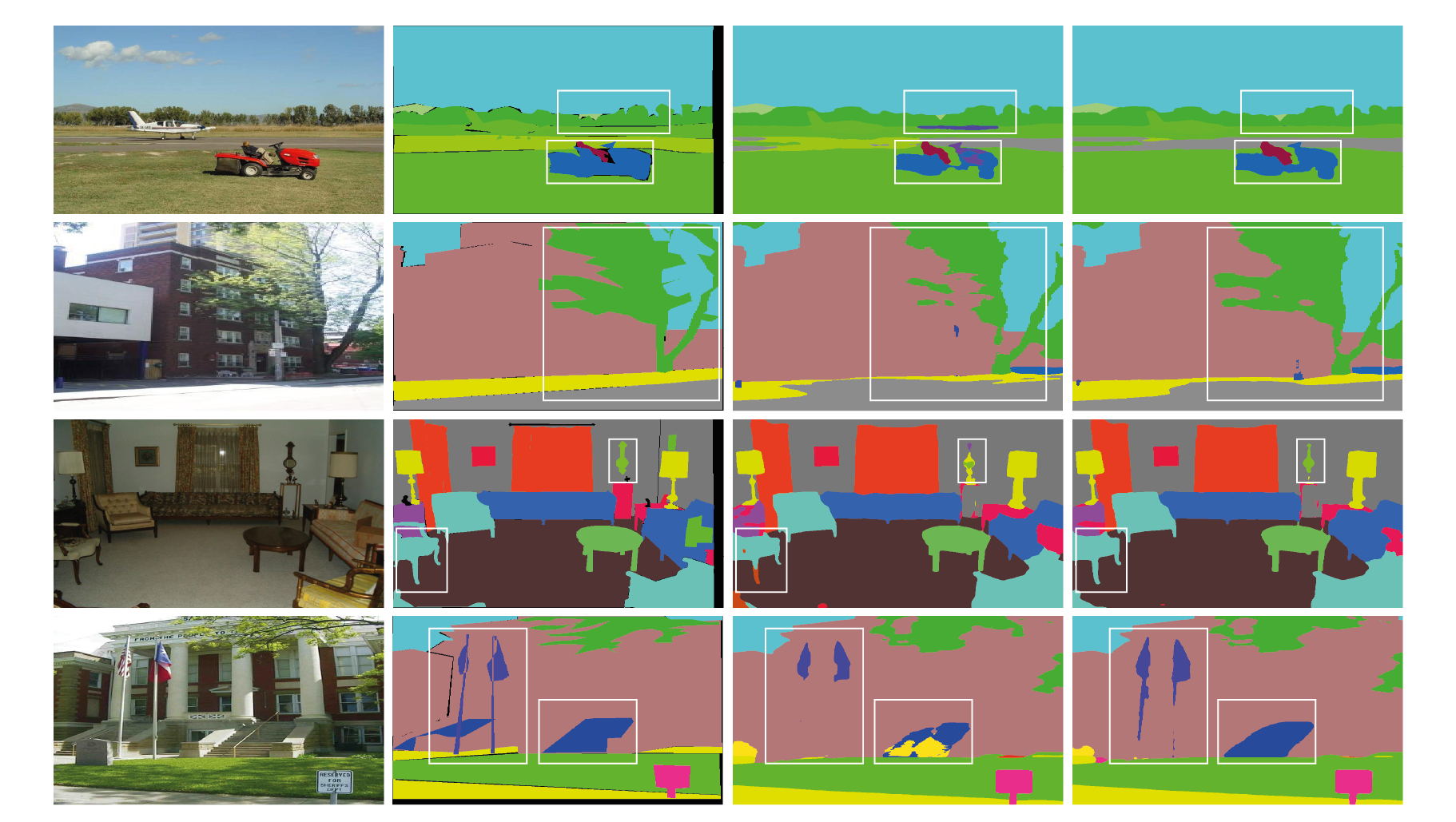}
\end{center}
\caption{\textbf{Qualitative comparisons on ADE20K val}: we compare UPerNet-Swin-S trained with only CE to the same model trained with also our multi- and cross-scale losses. White bounding boxes indicate some examples where our model performs better in cases where the foreground class is difficult to distinguish from the background (\nth{2}, \nth{5}  row) or when it recognizes and segments smaller/thinner objects missed by the baseline (\nth{1}, \nth{3}, \nth{6}, \nth{7})}.
\label{fig:ade20k_qualitative_supp}
\end{figure*}

\begin{figure*}
\begin{center}
\includegraphics[width=.99\linewidth]{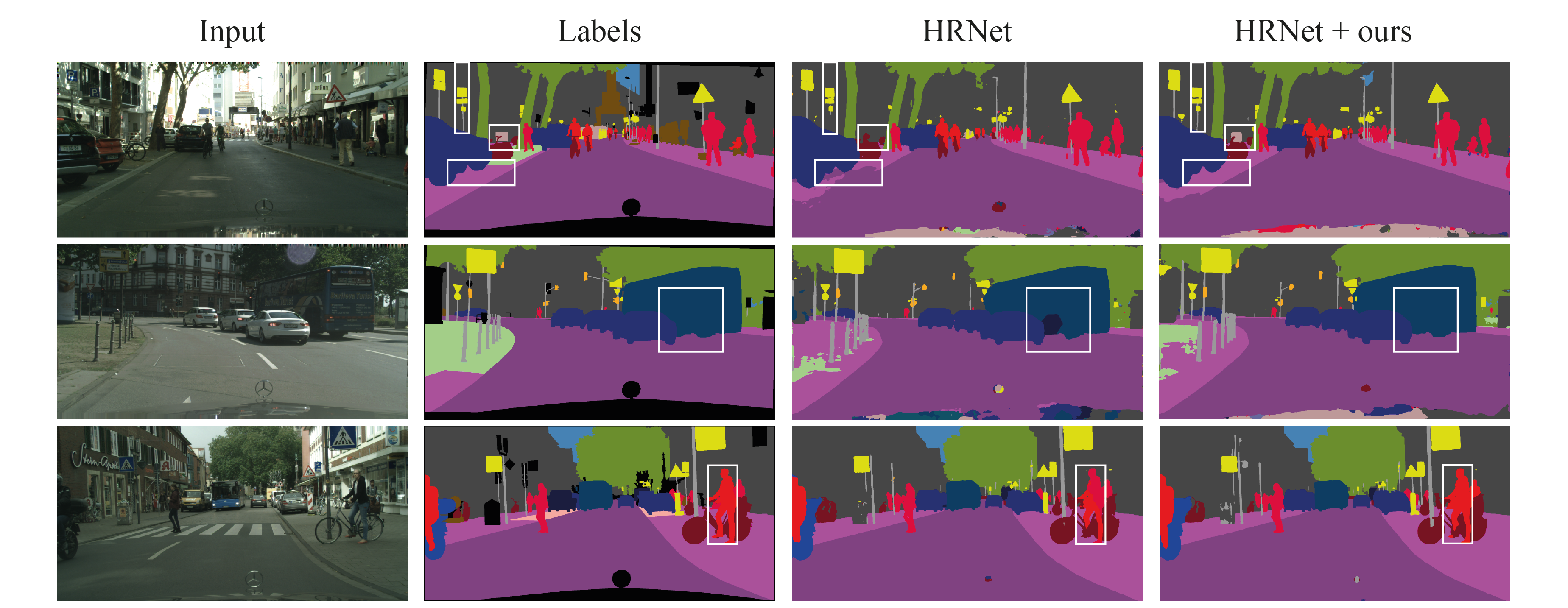}

\includegraphics[width=.92\linewidth]{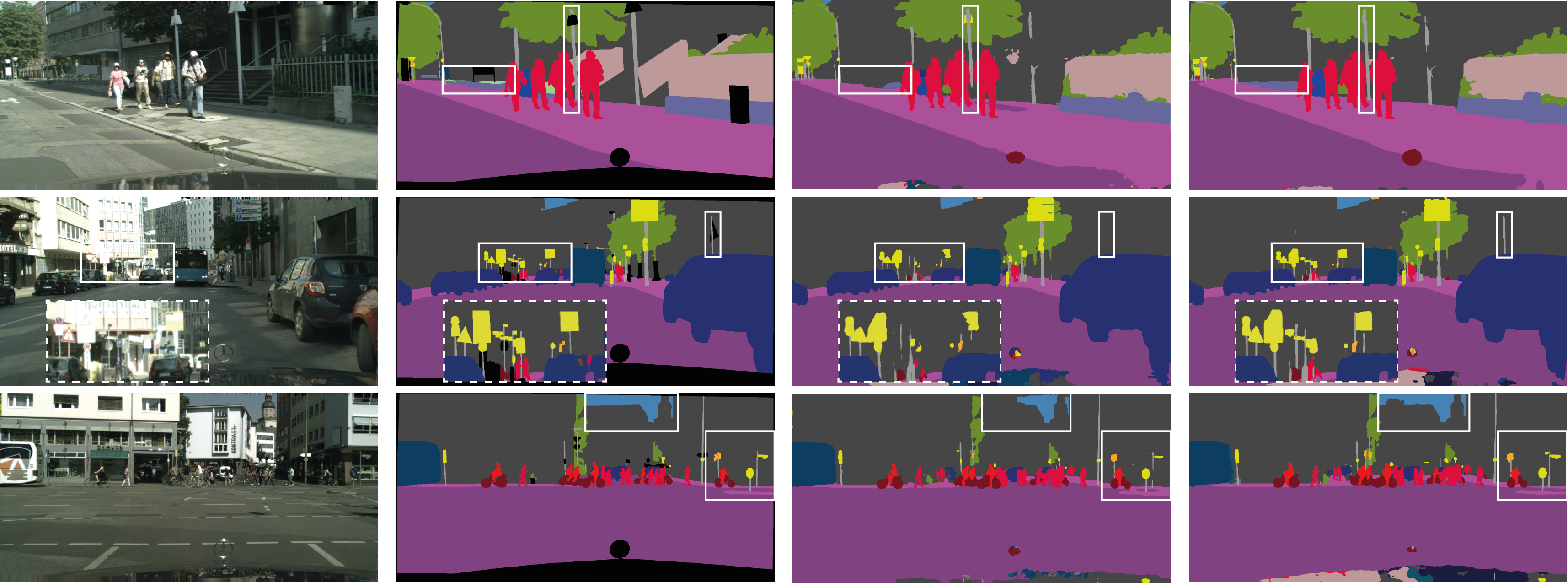}
\end{center}
\caption{\textbf{Qualitative results on Cityscapes validation set}: we present more qualitative results, comparing HRNet trained with CE to HRNet it trained with our multi- and cross-scale losses. White bounding boxes, outline some of the differences between the two models. Notably, the $\nth{2}$ and $\nth{3}$ rows depict cases where the baseline, misclassifies local segments of an object instance, namely a bus is partially recognized as "truck" and a bike rider is partially recognized as a simple pedestrian (i.e "person" in the dataset classes). Our model does not produce these inconsistencies in those cases, showcasing better ability to consider local-global interactions in recognizing and delineating an object instance. Other rows demonstrate examples where the model trained with our loss performs better than the baseline, at delineating small objects with fine details such as traffic signs or poles.}

\label{fig:cts_qualitative_supp}
\end{figure*}

\begin{figure*}
\begin{center}
\includegraphics[width=.99\linewidth]{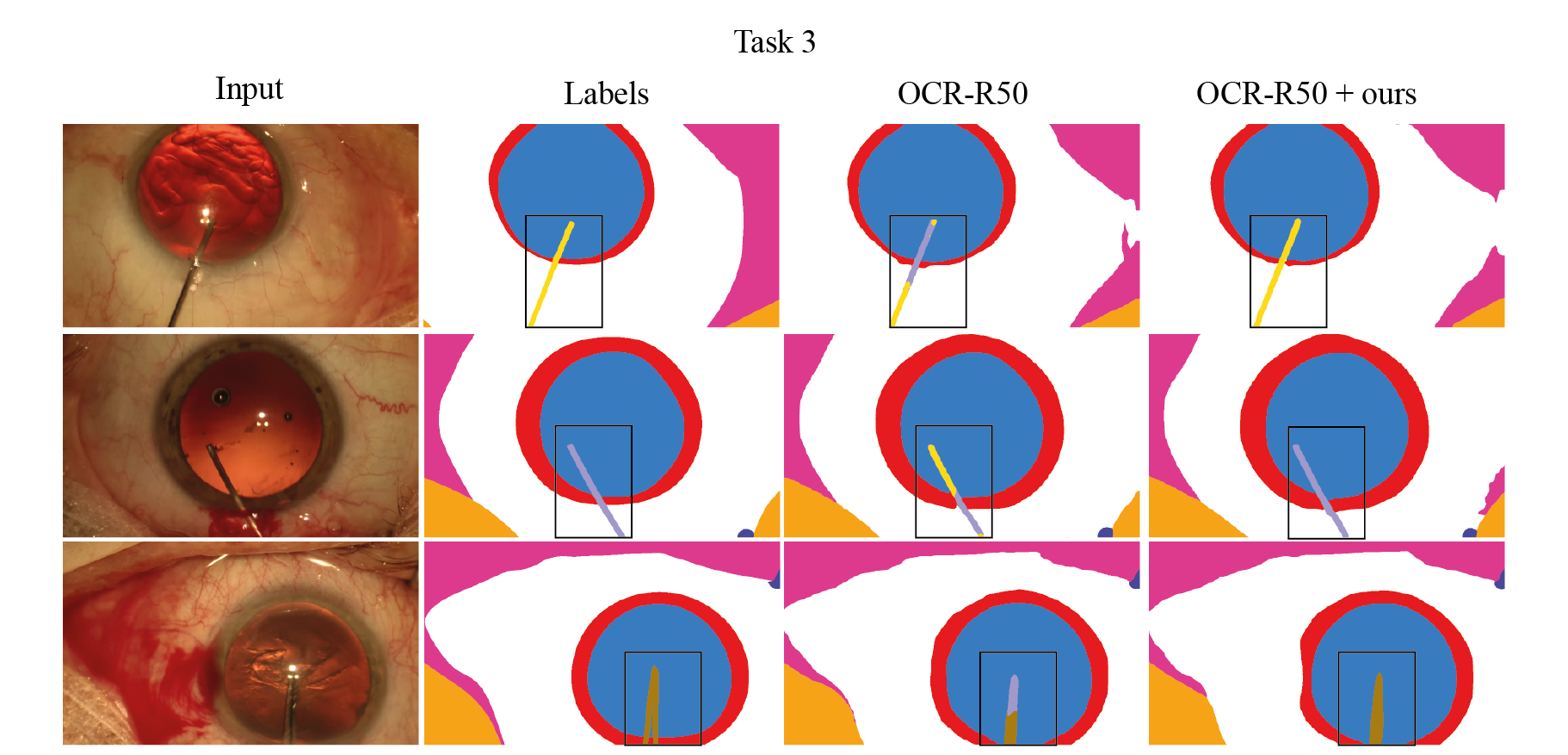}

\vfill
\includegraphics[width=.99\linewidth]{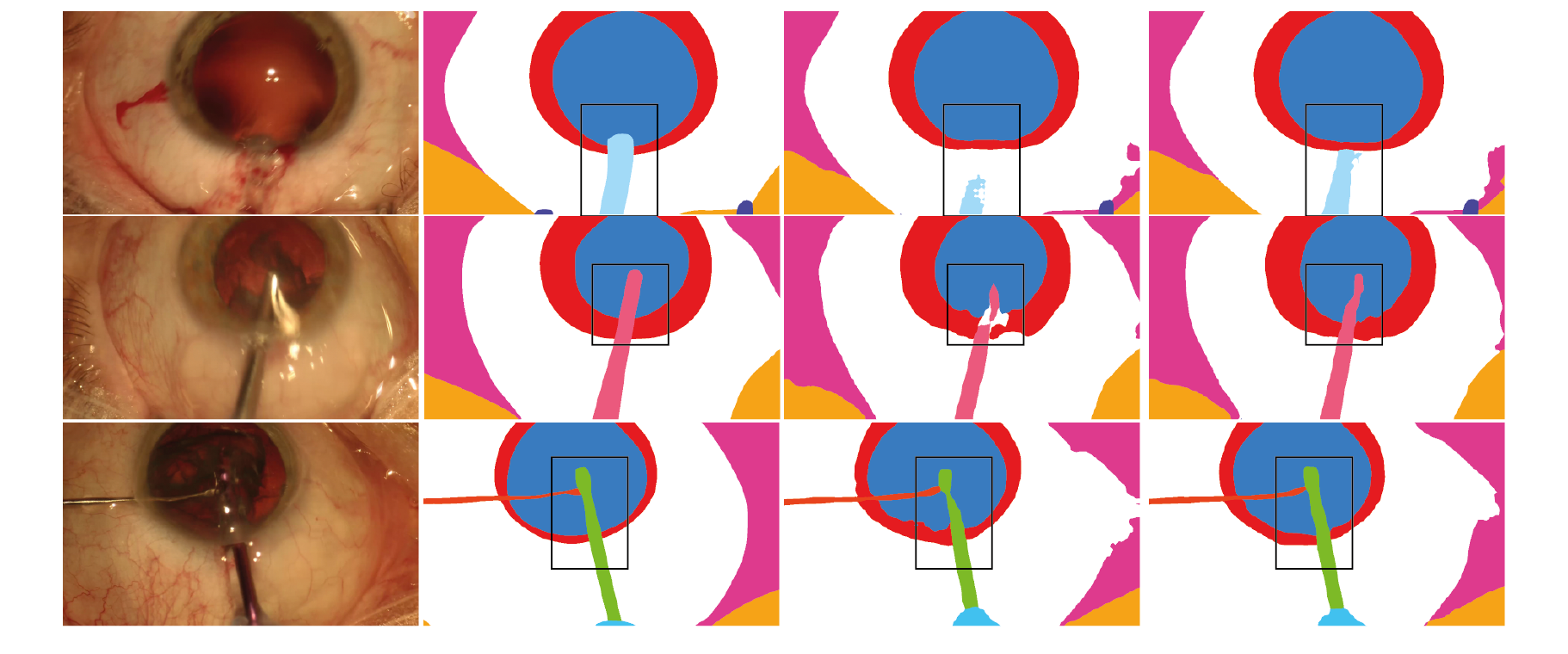}
\end{center}

\caption{\textbf{Qualitative results on CaDIS validation set for task $3$}: We present a visual comparison of the baseline and the result of combining it with our multi- and cross-scale losses. Rows $1$-$3$ demonstrate falsely recognized instrument classes by the baseline whereas our result accurately segments and classifies the tools. Notably, all $3$ cases, correspond to tools that have very similar appearance but should be discriminated in task $3$, that requires fine grained segmentation and classification. Further, rows $4$-$6$ demonstrate, a barely humanly visible translucent tool and two blurry and specular images with tools, respectively. In all three cases our model achieves clearly more accurate delineation of the tools than the baseline, under challenging conditions.}
\label{fig:cadis_qualitative_supp}
\end{figure*}

We provide qualitative results of models trained with our proposed loss on ADE20K (Fig. \ref{fig:ade20k_qualitative_supp}), Cityscapes (Fig. \ref{fig:cts_qualitative_supp}) and CaDIS (Fig. \ref{fig:cadis_qualitative_supp}). We also compare the feature spaces of UPerNet with ResNet-$101$ backbone, without and with our contrastive loss (Fig. \ref{fig:tsne_supp}).


% \newpage

\section{Training and testing settings}
In Tables \ref{tab:settings_ade20k},\ \ref{tab:settings_cts}, \ref{tab:settings_pascal} and \ref{tab:settings_cadis} we provide the settings 
used for our experiments. We closely follow each baseline's implementation details found in its official code publication. 
Regarding testing, when multi-scale (flipping and scaling) inference is used, the scaling factors used are $0.5, 0.75, 1.25, 1.5, 1.75$ on 
\textbf{ADE20K} and $0.5, 0.75, 1.25, 1.5, 1.75, 2.0$ on \textbf{Cityscapes-test} and \textbf{Pascal-Context}. 

\begin{table*}[htb]
  \centering
    \caption{Training settings on \textbf{Cityscapes}.}

\resizebox{.9\linewidth}{!}{
  \begin{tabular}{*8c}
    \toprule
    \multicolumn{2}{c}{Network}  &  \multicolumn{6}{c}{Settings}                     \\
    \cmidrule(r){1-2}               \cmidrule(l){3-8}
    \tbf{Model}   & \tbf{Backbone}  & \textbf{crop}   & \textbf{lr} & \textbf{decay} & $w_{d}$   & \textbf{Batch/steps}    & \textbf{optim} \\
    \midrule
    \mc{\hrn}     & \mc{HR$48$v$2$} & $512\times1024$ & $10^{-2}$ & poly & $5\times10^{-5}$ & \mc{$12/120K$} & SGD \\ 
    \mc{OCRNet}   & \mc{HR$48$v$2$} & $512\times1024$ & $10^{-2}$ & poly & $5\times10^{-5}$ & \mc{$12/120K$} & SGD \\ 
    \mc{\dv}      & \mc{R$101$}     & $512\times1024$ & $10^{-2}$ & poly & $5\times10^{-5}$ & \mc{$12/120K$} & SGD \\ 
    \mc{UPerNet}  & \mc{R$101$}     & $512\times1024$ & $10^{-2}$ & poly & $5\times10^{-5}$ & \mc{$12/120K$} & SGD \\
    \mc{UPerNet}  & \mc{Swin-T}     & $512\times1024$ & $6\times10^{-5}$ & linear & $10^{-2}$ & \mc{$8/120K$} & ADAMW \\
    \mc{UPerNet}  & \mc{Swin-S}     & $512\times1024$ & $6\times10^{-5}$ & linear & $10^{-2}$ & \mc{$8/120K$} & ADAMW \\
    \mc{UPerNet}  & \mc{Swin-B}     & $512\times1024$ & $6\times10^{-5}$ & linear & $10^{-2}$ & \mc{$8/120K$} & ADAMW \\
    \bottomrule
  \end{tabular}
  } % resizebox  
  \label{tab:settings_cts} 
\end{table*}

\vspace{-15mm}

\begin{table*}[htb]
  \centering
    \caption{Training settings on \textbf{ADE20K}.}

\resizebox{.9\linewidth}{!}{
  \begin{tabular}{*8c}
    % \begin{tabular}{cc|c|c|c|c|c}
    \toprule
    \multicolumn{2}{c}{Network}  &  \multicolumn{6}{c}{Settings}                     \\
    \cmidrule(r){1-2}               \cmidrule(l){3-8}
    \tbf{Model}   & \tbf{Backbone}  & \textbf{crop}   & \textbf{lr} & \textbf{decay} & $w_{d}$   & \textbf{Batch/steps}    & \textbf{optim} \\
    \midrule
    \mc{OCRNet}   & \mc{HR$48$v$2$} & $512\times512$ & $10^{-2}$ & poly           & $10^{-4}$        & \mc{$16/160K$}  & SGD \\ 
    \mc{\dv}      & \mc{R$101$}     & $512\times512$ & $10^{-2}$ & poly           & $10^{-4}$        & \mc{$16/160K$}  & SGD \\ 
    \mc{UPerNet}  & \mc{R$101$}     & $512\times512$ & $10^{-2}$ & poly           & $10^{-4}$        & \mc{$16/160K$}  & SGD \\ 
    \mc{UPerNet}  & \mc{Swin-T}     & $512\times512$ & $6\times10^{-5}$ & linear   & $10^{-2}$        & \mc{$16/160K$}  & ADAMW \\
    \mc{UPerNet}  & \mc{Swin-S}     & $512\times512$ & $6\times10^{-5}$ & linear   & $10^{-2}$        & \mc{$16/160K$}  & ADAMW \\
    \mc{UPerNet}  & \mc{Swin-B}     & $512\times512$ & $6\times10^{-5}$ & linear   & $10^{-2}$        & \mc{$16/160K$}  & ADAMW \\
    \mc{UPerNet}  & \mc{Swin-L}     & $640\times640$ & $6\times10^{-5}$ & linear   & $10^{-2}$        & \mc{$16/160K$}  & ADAMW \\
    % ll||ccc|ccc|ccc||c|c
    \bottomrule
  \end{tabular}
  } % resizebox  
  \label{tab:settings_ade20k} 
\end{table*}

\vspace{-15mm}

\begin{table*}[htb]
  \centering
    \caption{Training settings on \textbf{Pascal-Context}.}

\resizebox{.9\linewidth}{!}{
  \begin{tabular}{*8c}
    % \begin{tabular}{cc|c|c|c|c|c}
    \toprule
    \multicolumn{2}{c}{Network}  &  \multicolumn{6}{c}{Settings}                     \\
    \cmidrule(r){1-2}               \cmidrule(l){3-8}
    \tbf{Model}   & \tbf{Backbone}  & \textbf{crop}   & \textbf{lr} & \textbf{decay} & $w_{d}$   & \textbf{Batch/steps}    & \textbf{optim} \\
    \midrule
    \mc{OCRNet}   & \mc{HR$48$v$2$} & $512\times512$ & $10^{-3}$ & poly           & $10^{-4}$        & \mc{$16/160K$}  & SGD \\ 
    \mc{\hrn}      & \mc{HR$48$v$2$}     & $512\times512$ & $10^{-3}$ & poly           & $10^{-4}$        & \mc{$16/160K$}  & SGD \\ 
    \bottomrule
  \end{tabular}
  } % resizebox  
  \label{tab:settings_pascal} 
\end{table*}

\vspace{-15mm}

\begin{table*}[htb]
  \centering
    \caption{Training settings on \textbf{CaDIS}.}

\resizebox{.9\linewidth}{!}{
  \begin{tabular}{*8c}
    % \begin{tabular}{cc|c|c|c|c|c}
    \toprule
    \multicolumn{2}{c}{Network}  &  \multicolumn{6}{c}{Settings}                     \\ 
    \cmidrule(r){1-2}               \cmidrule(l){3-8}
    \tbf{Model}   & \tbf{Backbone}  & \textbf{crop}   & \textbf{lr} & \textbf{decay} & $w_{d}$   & \textbf{Batch/steps}    & \textbf{optim} \\ 
    \midrule
    \mc{OCRNet}   & \mc{R$50$} & $540\times960$ & $10^{-4}$ & exp           & -        & \mc{$8/20K$}  & ADAM \\ 
    \bottomrule
  \end{tabular}
  } % resizebox  
  \label{tab:settings_cadis} 
\end{table*}

\vfill
\vfill

% \newpage
% \section{Pseudo-code}
% We provide the pseudo-code \ref{algo:pseudo} that describes the computation of our multi-scale and cross-scales losses. We also intend to make our code available online.
% \input{tab/pseudocode}

%%%%% sections %%%%%%

\clearpage
\bibliographystyle{splncs04}
\bibliography{macros,main}
% \newpage
% \input{sec/X_supplementary}

\end{document}